\def\year{2023}\relax
\documentclass[letterpaper]{article} % DO NOT CHANGE THIS
\usepackage{aaai23}  % DO NOT CHANGE THIS
\usepackage{times}  % DO NOT CHANGE THIS
\usepackage{helvet}  % DO NOT CHANGE THIS
\usepackage{courier}  % DO NOT CHANGE THIS
\usepackage[hyphens]{url}  % DO NOT CHANGE THIS
\usepackage{graphicx} % DO NOT CHANGE THIS
\usepackage{natbib}  % DO NOT CHANGE THIS AND DO NOT ADD ANY OPTIONS TO IT
\usepackage{caption} % DO NOT CHANGE THIS AND DO NOT ADD ANY OPTIONS TO IT
\DeclareCaptionStyle{ruled}{labelfont=normalfont,labelsep=colon,strut=off} % DO NOT CHANGE THIS
\usepackage{algorithm}
\usepackage{algorithmic}

\usepackage{newfloat}
\usepackage{listings}
\floatstyle{ruled}
\newfloat{listing}{tb}{lst}{}
\floatname{listing}{Listing}
\usepackage{latexsym}
\usepackage{comment}

\usepackage{color} % 最終的にはNG
\usepackage{multirow} %ok
\usepackage{amsmath, amsfonts, amsthm, amssymb}
\usepackage{subfigure} %ok
\usepackage[inline]{enumitem}
\usepackage{booktabs}

\usepackage[switch]{lineno} % camera readyで消す

\newtheoremstyle{mysubtask}%
{}{}{\normalfont}{}{\scshape}{.\,}{ }{}
\theoremstyle{mysubtask}
\newtheorem{subtask}{Subtask}

\newtheoremstyle{mytask}%
{}{}{\normalfont}{}{\scshape}{.\,}{ }{}
\theoremstyle{mytask}

\newtheoremstyle{mydef}%
{}{}{\normalfont}{}{\scshape}{.\,}{ }{}
\theoremstyle{mydef}

\date{}

\title{SlideVQA: A Dataset for Document Visual Question Answering on Multiple Images}
\author{Anonymous Submission}

\author{
    Ryota Tanaka,
    Kyosuke Nishida, 
    Kosuke Nishida,
    Taku Hasegawa,
    Itsumi Saito,
    Kuniko Saito\\
}
\affiliations{
    NTT Human Informatics Laboratories, NTT Corporation \\
    \{ryouta.tanaka.rg, kyosuke.nishida.rx, kosuke.nishida.ap, taku.hasegawa.ps, itsumi.saito.df, kuniko.saito.ku\}@hco.ntt.co.jp
}

\begin{document}
%\linenumbers

\maketitle

\begin{abstract}
Visual question answering on document images that contain textual, visual, and layout information, called document VQA, has received much attention recently. Although many datasets have been proposed for developing document VQA systems, most of the existing datasets focus on understanding the content relationships within a single image and not across multiple images. In this study, we propose a new multi-image document VQA dataset, SlideVQA, containing 2.6k+ slide decks composed of 52k+ slide images and 14.5k questions about a slide deck. SlideVQA requires complex reasoning, including single-hop, multi-hop, and numerical reasoning, and also provides annotated arithmetic expressions of numerical answers for enhancing the ability of numerical reasoning. Moreover, we developed a new end-to-end document VQA model that treats evidence selection and question answering in a unified sequence-to-sequence format. Experiments on SlideVQA show that our model outperformed existing state-of-the-art QA models, but that it still has a large gap behind human performance. We believe that our dataset will facilitate research on document VQA.
\end{abstract}

\section{Introduction}

Building intelligent agents that can read and comprehend real-world documents, such as webpages, office documents, lecture slides, etc., has been a long-standing goal of artificial intelligence. To achieve this goal, machine reading comprehension (MRC), a central task in natural language understanding, has been intensively studied.
The typical definition of the MRC task is quite simple, wherein given a short natural language text as a context and a question about it, a machine reads the text and then answers the question by extracting a span from the text~\cite{RajpurkarZLL16,RajpurkarJL18}. However, this definition is far from real-world applications, such as customer service chatbots on e-commerce websites~\cite{CuiHWTDZ17} and assistant systems for reading professional literature~\cite{HongWJZW19},
in that the context is composed entirely of text, with no graphical elements. 

To this end, visual question answering on document images (document VQA) has received much attention. It is a challenging vision and language task that requires methods to reason about document layout, textual content, and visual elements~\cite{Mathew_2021_WACV,DBLP:conf/aaai/TanakaNY21,Mathew_2022_WACV}. When the primary content in a document is text (e.g., e-mails and forms) 
and the task is to understand it on the basis of its layout information, state-of-the-art models have already achieved 
nearly human-level performance~\cite{xu2020layoutlmv2,powalski2021going}. 
On the other hand, challenges remain when it comes to handling diverse real-world documents.
First and foremost is that current models are not capable of performing reasoning across 
multiple images since the existing datasets focus on testing reasoning ability on a single image.
Moreover, compared with humans, document VQA models still have trouble understanding documents that contain visual elements and understanding questions that require numerical reasoning~\cite{Mathew_2022_WACV}.

\begin{figure*}[t!]
    \centering
    \includegraphics[width=.99\textwidth]{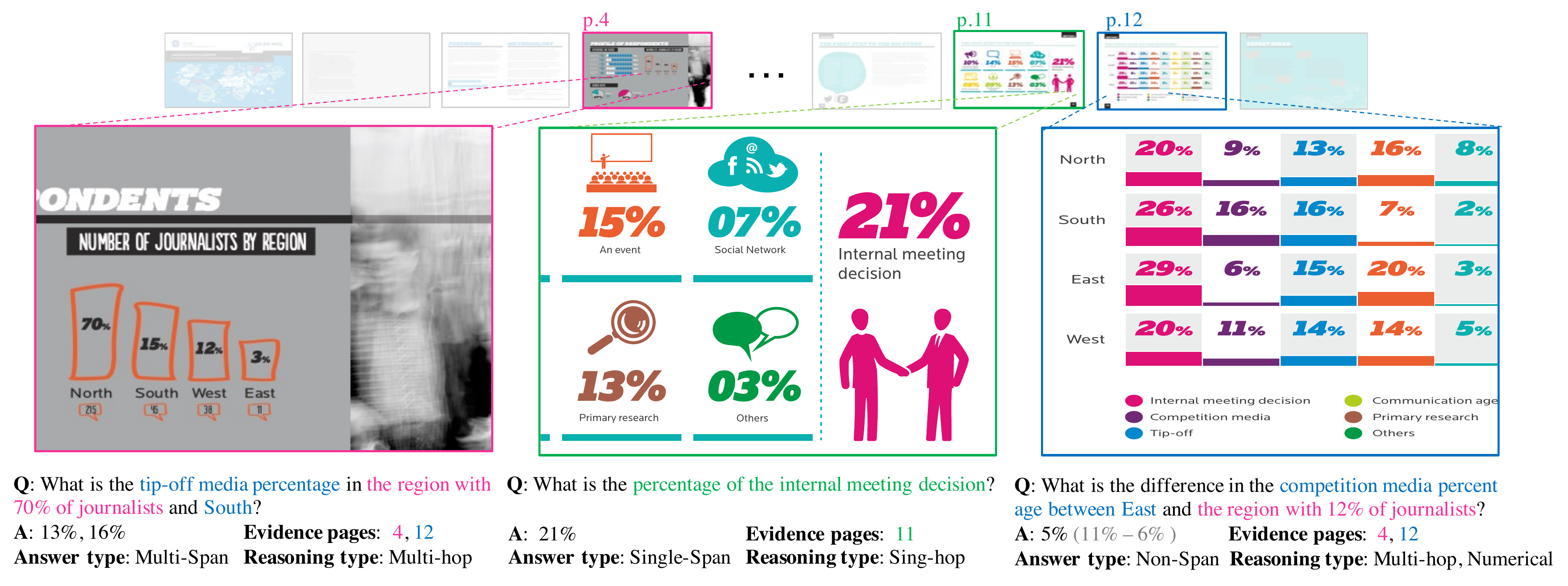}
    \caption{Examples from our SlideVQA dataset. Some questions can be answered through single-hop, multi-hop, and numerical reasoning. The colors of the words match the image borders with the same colors. ($\cdot$) of the right example in the answer denotes an annotated arithmetic expression to derive the final answer. The slide deck can be viewed at~\url{https://www.slideshare.net/mslgroup/mediainsights-evolving-sources-of-news-for-media}.}
\label{fig:example_dataset}
\end{figure*}

To address the above challenges, we introduce a new document VQA dataset\footnote{Our dataset and codes are publicly available at~\url{https://github.com/nttmdlab-nlp/SlideVQA}}, SlideVQA, for tasks wherein given a slide deck composed of multiple 
slide images
and a corresponding question, a system selects a set of evidence images and answers the question. Slide decks are one of the most efficient document types that arrange visual and textual elements for communication. As shown in Figure~\ref{fig:example_dataset}, SlideVQA requires complex reasoning over slide images, including single-hop, multi-hop, and numerical reasoning. These reasoning skills play essential roles in MRC tasks~\cite{Yang0ZBCSM18,dua-etal-2019-drop}.

Our main contributions are summarized as follows:
\begin{itemize}
\item We introduce a novel task and dataset,
SlideVQA, wherein to answer its questions, a machine has to read and comprehend a slide deck.
It is the largest multi-image document VQA dataset containing 2.6k+ slide decks (each consisting of 20 slides) and 14.5k questions. It also provides bounding boxes around textual and visual elements for understanding document layout and arithmetic expressions for numerical reasoning.

\item  We developed a  
\textbf{M}ulti-\textbf{M}odal \textbf{M}ulti-image \textbf{D}ocument VQA model, M3D, to jointly
perform evidence selection and question answering tasks
and to enhance numerical reasoning by generating arithmetic expressions. 
\item Our model outperformed existing state-of-the-art QA models on SlideVQA, but its performance is still below that of humans by a large margin.
\end{itemize}

\section{Related Work}
\subsubsection{Datasets for VQA on document images.}
Document VQA is the task of answering questions about %real-world 
document images, and 
some useful 
datasets have been published, such as DocVQA~\cite{Mathew_2021_WACV},
VisualMRC~\cite{DBLP:conf/aaai/TanakaNY21},
WebSRC~\cite{ChenZCJZLX021}, and
InfographicVQA~\cite{Mathew_2022_WACV}.
The task assumes that the datasets have a single relevant %document
image, containing all the facts required to answer. % the question. %, is given for answering the question.

The work most related to ours is DocCVQA~\cite{tito2021document}, wherein a large collection of document images is used to answer a given question. Our dataset differs from DocCVQA, as follows. First, SlideVQA 
%is the largest multi-image document VQA dataset consisting
consists
of 14.5k questions,
%\textcolor{magenta}{based on a slide deck}, %based on multiple slides, 
% slide (document) images
wheres DocCVQA provides only 20 questions. Second, SlideVQA requires multi-hop reasoning over multiple %\textcolor{blue}{document images}
slides
to find the answer, while DocCVQA requires %\textcolor{blue}{and other datasets require} 
only single-hop reasoning on individual images to find the answer.
Besides these differences, SlideVQA provides questions that require numerical reasoning and arithmetic expression annotations to answer numerical questions (e.g., ``30 - 28" for the answer ``2"): no other VQA dataset, including InfographicVQA that requires numerical reasoning, provides such annotations. Furthermore, SlideVQA provides the largest number of bounding boxes on all of the collected images among the related datasets. %This will facilitate the development of document layout understanding. 
%; (iii) SlideVQA provides questions that require numerical reasoning and arithmetic expression annotations to answer the question (e.g., ``30 - 28" for answer ``2"): no other dataset, including InfographicVQA that requires numerical reasoning, provides such annotations. Besides these differences, SlideVQA provides the largest number of bounding boxes on all of the collected images of any of the related datasets. This will facilitate the development of document layout understanding. 

%Compared with InfographicVQA~\cite{Mathew_2022_WACV}, which requires numerical reasoning, SlideVQA additionally provides arithmetic expressions as intermediate reasoning steps to answer numerical questions for improving numerical reasoning.

\begin{table*}[t!]
    \centering
       \scalebox{0.825}{
\small
    \tabcolsep=3pt    
    \begin{tabular}{ccccccccccccc} 
    \toprule
         \multirow{2}{*}{Dataset}& Document &  Multi-images & Multi-hop & Numerical &  Answer & Document images &  \multirow{2}{*}{\#QAs} &  \multirow{2}{*}{\#Images} & \multirow{2}{*}{\#BBoxes} & \#Arithmetic & \#Evidence \\
        & source & input & reasoning & reasoning & type & modal type &  &  & & annotations & candidates \\ 
        \midrule
        DocVQA & industry  & & & & SS & TL & 50k & 12k  & -- & -- & 1\\
        VisualMRC & web-pages & & & & Ab & TLV & 30k & 10k & 64k & -- & 1\\
        WebSRC & web-pages & & & & SS & TLV & 400k & 6.4k & -- & -- & 1\\
        InfographicVQA & infographics & & & \checkmark & SS, MS, NS & TLV & 30k & 5k & -- & -- & 1\\ 
        DocCVQA & industry  & \checkmark  & && MS & TL & 0.02k & 14k & -- & -- & 14k \\ 
        \midrule
        SlideVQA (Ours) & slide decks & \checkmark & \checkmark & \checkmark & SS, MS, NS & TLV & 14.5k & 52k & 890k & 1.7k & 20\\  
        \bottomrule
    \end{tabular}
    }
    \caption{Comparison of question answering datasets on document images. Answer types can be broken down into abstractive (Ab), single-span (SS), multi-span (MS), and non-span (NS). ``T/L/V'' denotes the ``text/layout/visual" modality of images.}
    \label{tab:statistics_dataset}
\end{table*}

\subsubsection{Document VQA Models.}
In parallel with the development of datasets, Transformer~\cite{VaswaniSPUJGKP17} has come to be used for understanding unstructured text in document images. LayoutLM~\cite{XuLCHWZ20}, LayoutLMv2~\cite{xu2020layoutlmv2}, LayoutT5~\cite{DBLP:conf/aaai/TanakaNY21}, and TILT~\cite{powalski2021going} have achieved impressive results in single-image document
%single-document 
VQA tasks by combining textual, layout, and visual features. By contrast, we focus on endowing models with the ability to reason and comprehend 
%multi-document images. 
multiple images.
Moreover, while \citet{tito2021document} used a pipeline  of retrieval and reading models for DocCVQA, we use multi-task learning that jointly performs evidence selection and question answering.

\subsubsection{Multi-modal question answering.}

%Multi-modal question answering
This type takes textual and visual information as input contexts, which is different from document VQA that takes only a document image as the input context.
%MovieQA~\cite{TapaswiZSTUF16} evaluates automatic story comprehension from both video and text.
%COMICS~\cite{IyyerMGVBDD17} test understanding of closure, transitions in the narrative from one panel of comic books to the next.
TQA~\cite{KembhaviSSCFH17} is comprised of middle-school science lessons containing diagrams and text.
%RecipeQA~\cite{YagciogluEEI18} provides cooking recipes with images and text.
MultiModalQA~\cite{talmor2021multimodalqa}  requires joint reasoning over text, tables, and images in Wikipedia.  
%The motivation behind these studies is similar to ours, but %their input is well-formed for machines, and the visual information in the text such as the document layout is dropped from the text in these datasets. 

\subsubsection{VQA on videos or image sets.} 
VideoQA focuses on answering questions about video frames of TV shows~\cite{lei-etal-2018-tvqa,lei-etal-2020-tvqa} and movies~\cite{tapaswi2016movieqa}. A similar task is VQA on image sets (ISVQA), which involves handling photos taken from different viewpoint indoors~\cite{bansal2020visual}. By contrast, our dataset also requires a model to understand the text in images.

%

%The focus of our research is to enable machines to handle the same visual input as humans do when they read real-world documents.

%multimodalQAはwikipediaに埋め込まれたテキスト・画像・テーブルをgivenで取得可能だが，我々の研究は文書画像を対象としており，文書中の要素を解析し，理解する必要がある

\subsubsection{Slide images understanding.} %Slide decks are one of the most efficient document types with which to exchange ideas on the Web, within educational institutes, and between businesses.
\citet{haurilet2019spase,haurilet2019wise} introduced a benchmark for object segmentation on slide-pages. 
%SlideVQA is the first object detection dataset to provide bounding box annotations for understanding slide images.
\citet{sun-etal-2021-d2s,fu2021doc2ppt} tackled the task of generating slides from research papers. 
Our work is the first to focus on answering questions on sets of slide images.

\subsubsection{Reasoning over textual documents.} Numerical reasoning plays an important role in NLP tasks~\cite{dua-etal-2019-drop,zhang-etal-2020-language,zhang-etal-2021-noahqa-numerical}. Moreover, multi-hop reasoning has taken the spotlight as it aligns with the multi-hop nature of how humans reason to acquire knowledge, and has led to a proliferation of benchmarks~\cite{talmor-berant-2018-web,Yang0ZBCSM18}. However, there is as yet no dataset for developing models to perform both multi-hop and numerical reasoning on document images. %Additionally, we focus on creating models with the ability to generate an arithmetic expression as an intermediate reasoning step that leads to the final answer for numerical reasoning. Chain of thought prompting~\cite{wei2022chain} improves the ability of large language models to perform complex reasoning such as math word problems by providing reasoning steps via prompting. Our work challenges the model to generate the arithmetic expressions itself rather than providing the expression as prompts by users.

\section{The SlideVQA Task and Dataset}

\subsection{Task Overview and Formulation}
The SlideVQA task, requires a system to answer a question about a slide deck, which is composed of an ordered set of slide images and to select evidence slide images. %\textcolor{blue}{Here, each image has textual and visual elements obtained by the OCR system and object detector, respectively.} 
We formulate the end-to-end SlideVQA task as follows: 
%\begin{task}[SlideVQA] 
\newline
\textsc{MainTask} (SlideVQA). 
\label{prob:main}
Given a question $q$ and a slide deck $\mathbf{I} = \{I_1, \ldots, I_{K}\}$ ($K=20$), 
a model outputs an answer $y$ and selects relevant slides $\mathbf{\hat{I}} = \{\hat{I}_1, \ldots, \hat{I}_{K'}\}$. 
%\end{task} 

The task can be decomposed into two subtasks: %the pipeline of two-task:
\begin{subtask}[Evidence Selection] 
\label{prob:es}
Given a question $q$ and a slide deck $\mathbf{I}$, a model identifies the images $\mathbf{\hat{I}}$ from which to derive the answer $y$. 
\end{subtask}
\begin{subtask}[Question  Answering] 
\label{prob:qa}
Given a question $q$ and 
%a slide deck $\mathbf{I}$ (or the selected slides $\mathbf{\hat{I}}$), 
%a set of 
the slide images ($\mathbf{I}$ or $\mathbf{\hat{I}}$),
a model outputs an answer $y$.
\end{subtask}

%The task 
SlideVQA has three answer types (see the examples in Figure~\ref{fig:example_dataset}). A single-span answer is a contiguous sequence of tokens in the reading order extracted from the image, and a multi-span answer is formed from multiple spans from the image. A non-span answer is not extracted and is composed of numerical values and visual appearances. 

We can also use annotations of bounding boxes around the objects (and their categories) to understand the semantic structure of images and annotations of arithmetic expressions to understand numerical reasoning as additional input at training. These annotations are not given at inference.

%\textcolor{magenta}{Here, this task assumes that each image has textual and visual elements obtained by the OCR system and object detector, respectively.} 
%in the training set, but these annotations are not given in the development and test sets.} 

\subsection{Dataset Collection}
In this section, we describe the collection process of the SlideVQA dataset. To control the annotation quality, we recruited crowd workers located in English-speaking countries and who had passed a rigorous qualification procedure. Additionally, we asked other workers to assess the quality of the annotated samples after each collection step. 

\subsubsection{Slide decks collection.}
First, %we used web crawlers to select and download 
we selected and downloaded
25,327 slide decks composed of more than 20 slides from slideshare\footnote{\url{https://www.slideshare.net/}} 
and covering 39 topics. We kept the first 20 slides and truncated the rest of the pages. Then, the workers filtered the collected decks that did not meet the following criteria: (i) the main language is English; (ii) the content is easy for workers to understand; (iii) the decks must contain one or more graphs, tables, figures, or numerical data to avoid creating questions requiring only text-level understanding. %After a verification process, we obtained a total of 2,619 slide decks.

\subsubsection{Bounding boxes and categories annotation.}
\begin{figure}[t!]
    \centering
    \includegraphics[width=.34\textwidth]{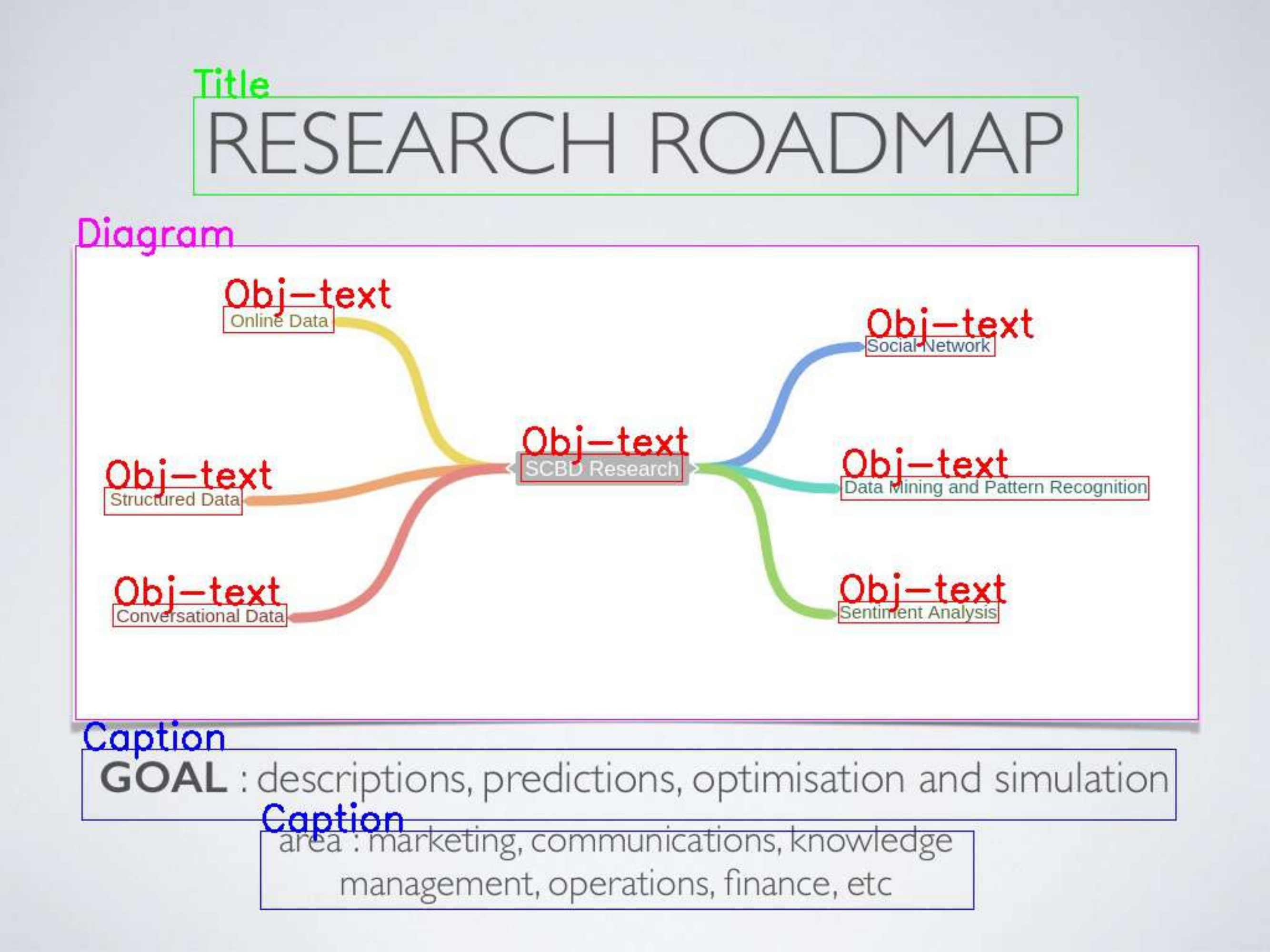}
    \caption{Example of collected bounding boxes. Colored boxes and words were annotated by workers. The image can be viewed at~\url{https://www.slideshare.net/andrybrewok/big-data-analytics-a-social-network-approach}.}
    \label{fig:bbox}
\end{figure}

To facilitate understanding of the semantic components of images, we annotated all images with bounding boxes and their categories. The workers indicated specific objects in each image by annotating bounding boxes around the objects and classifying them into nine classes that were based on SPaSe~\cite{haurilet2019spase} as follows:

\begin{itemize}
    \item \textbf{Title}: presentation title, slide title
    \item \textbf{Page-text}: text in slide, bullet-point text list, text list
    \item \textbf{Obj-text}: text in a figure, image, diagram or table
    \item \textbf{Caption}: description of figure, image, diagram, or table
    \item \textbf{Other-text}: footnote, date, affiliation, code, URL
    \item \textbf{Diagram}: a graphical representation of data, a process
    \item \textbf{Table}: data arranged in rows and columns
    \item \textbf{Image}: drawing, logo, map, screenshot, realistic image
    \item \textbf{Figure}: graph with data points and coordinates
\end{itemize}
As shown in Figure~\ref{fig:bbox}, SlideVQA provides densely annotated bounding boxes in images. 

\begin{figure}[t!]
    \centering
    \subfigure[Bounding box categories.]{
        \includegraphics[width=.22\textwidth]{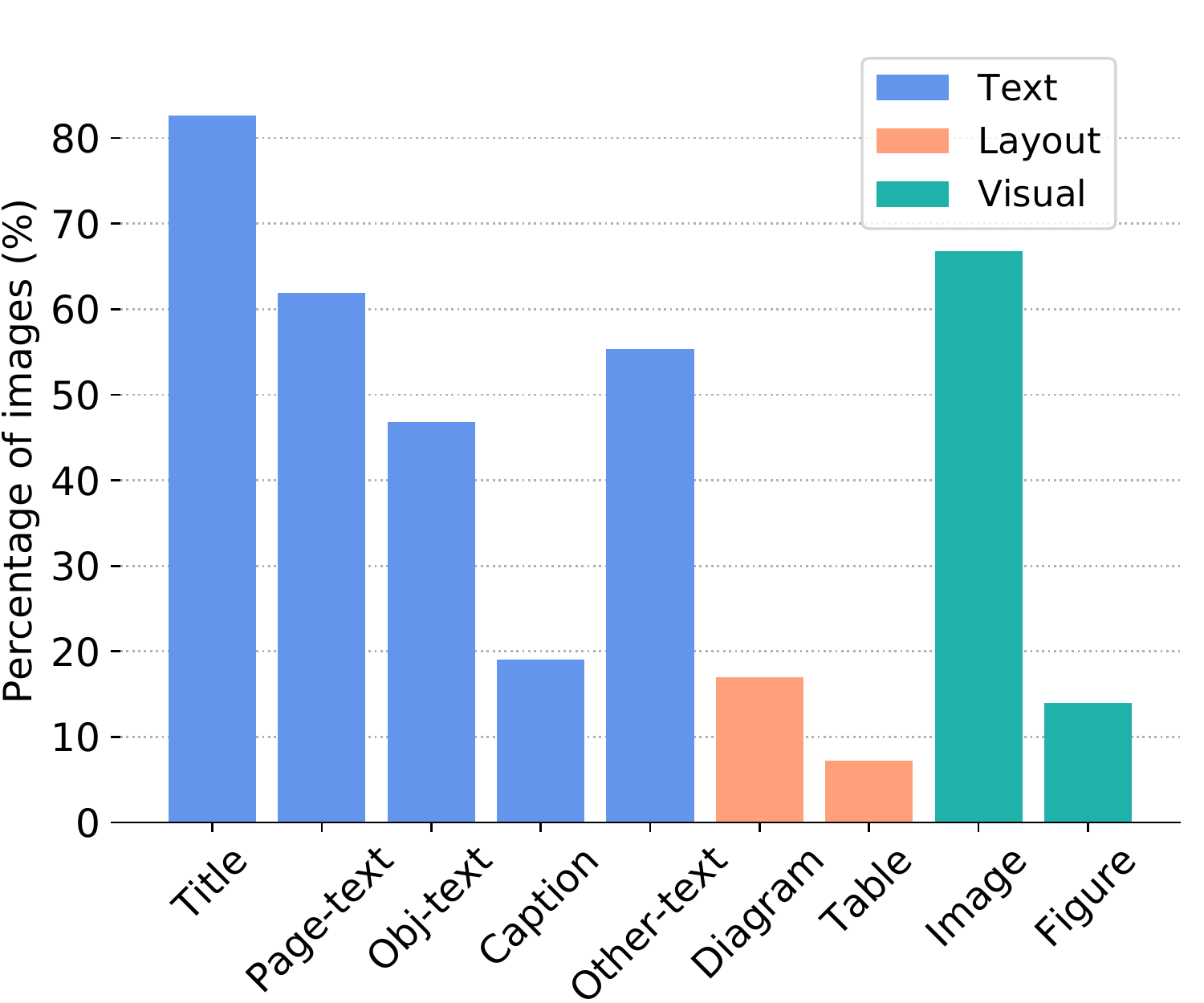}
    }    
    \subfigure[Reasoning types.]{
        \includegraphics[width=.22\textwidth]{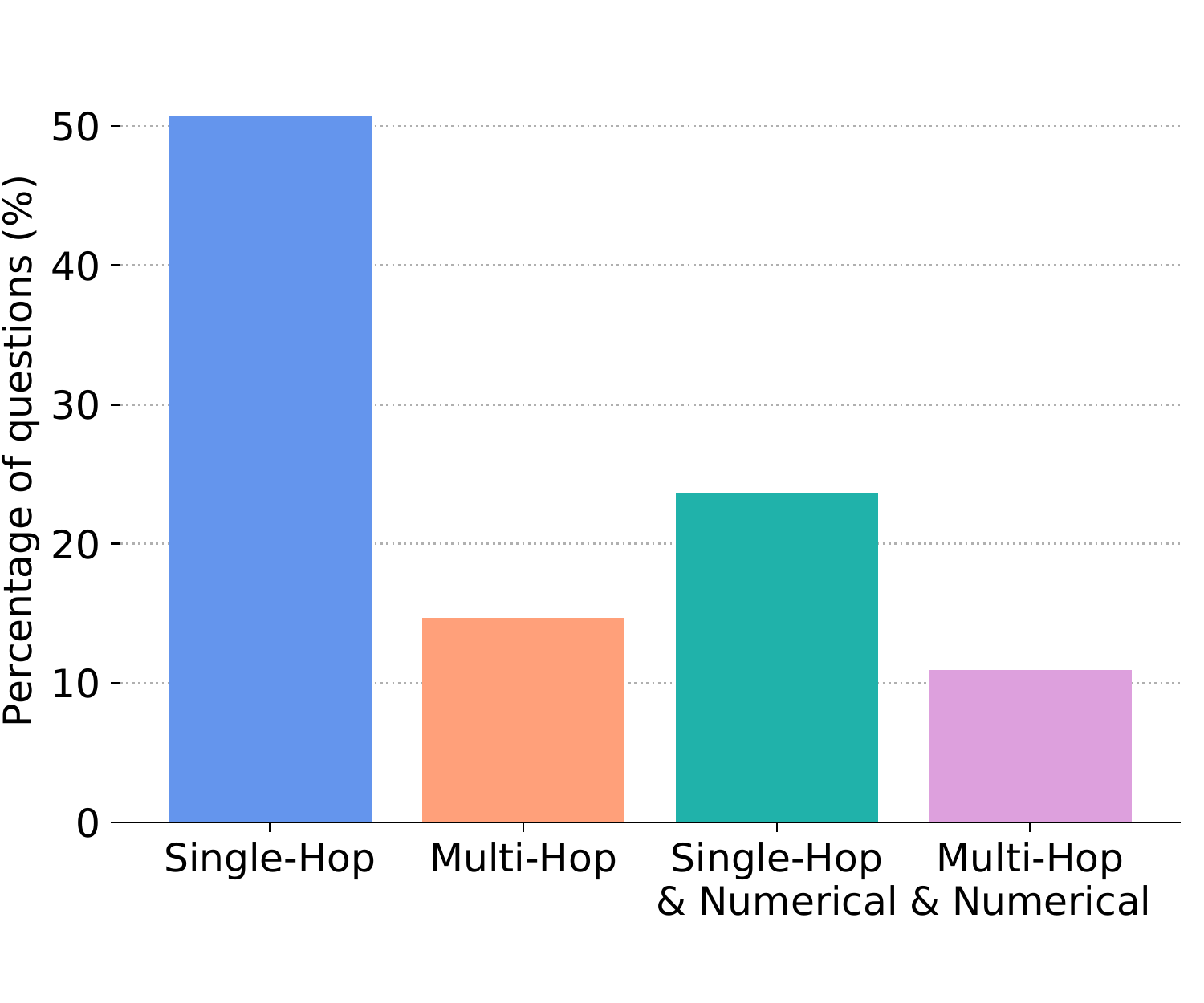}
    } \\
    \subfigure[Numerical operation types.]{
        \includegraphics[width=.22\textwidth]{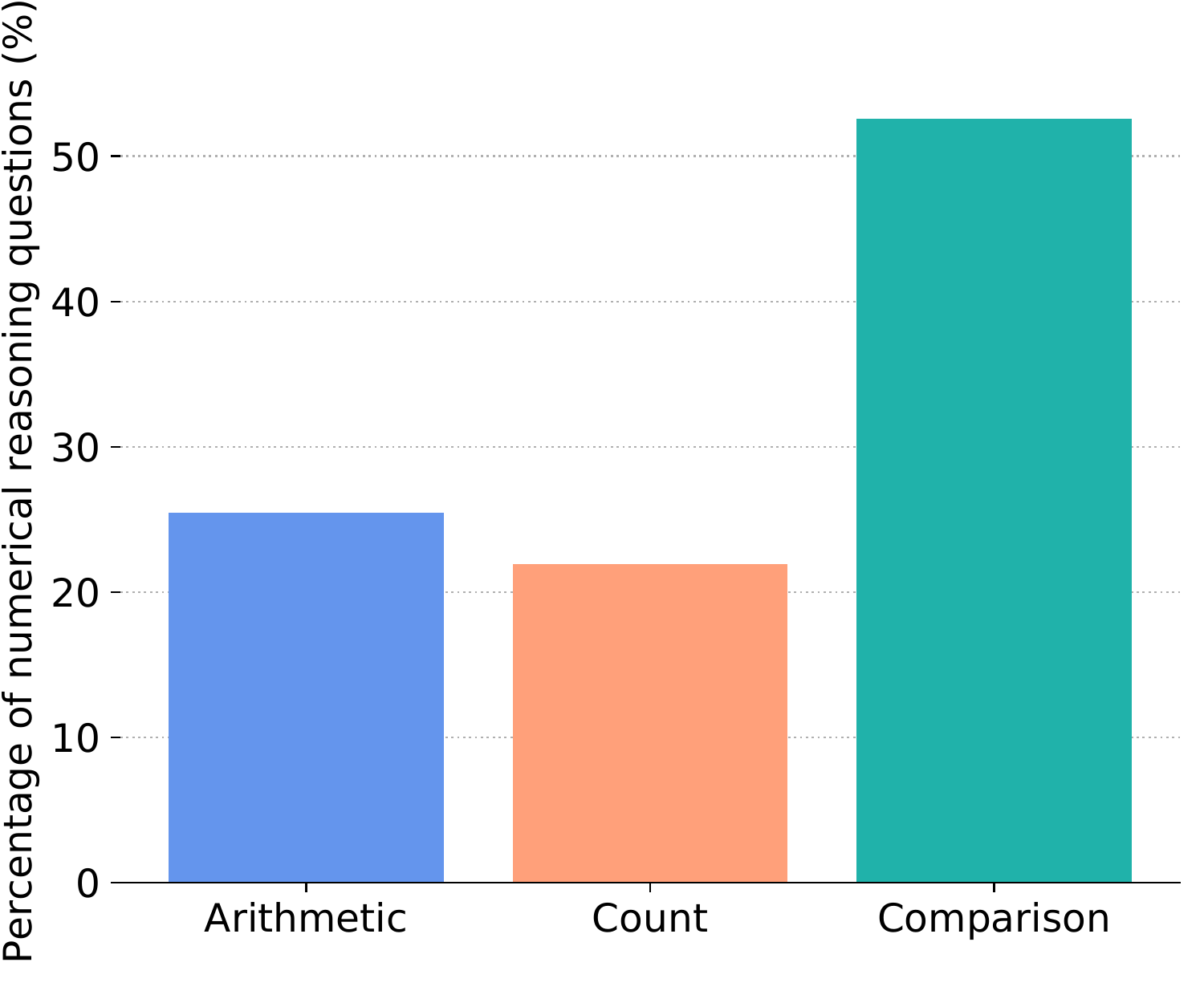}
    }
    \subfigure[Answer types.]{
        \includegraphics[width=.22\textwidth]{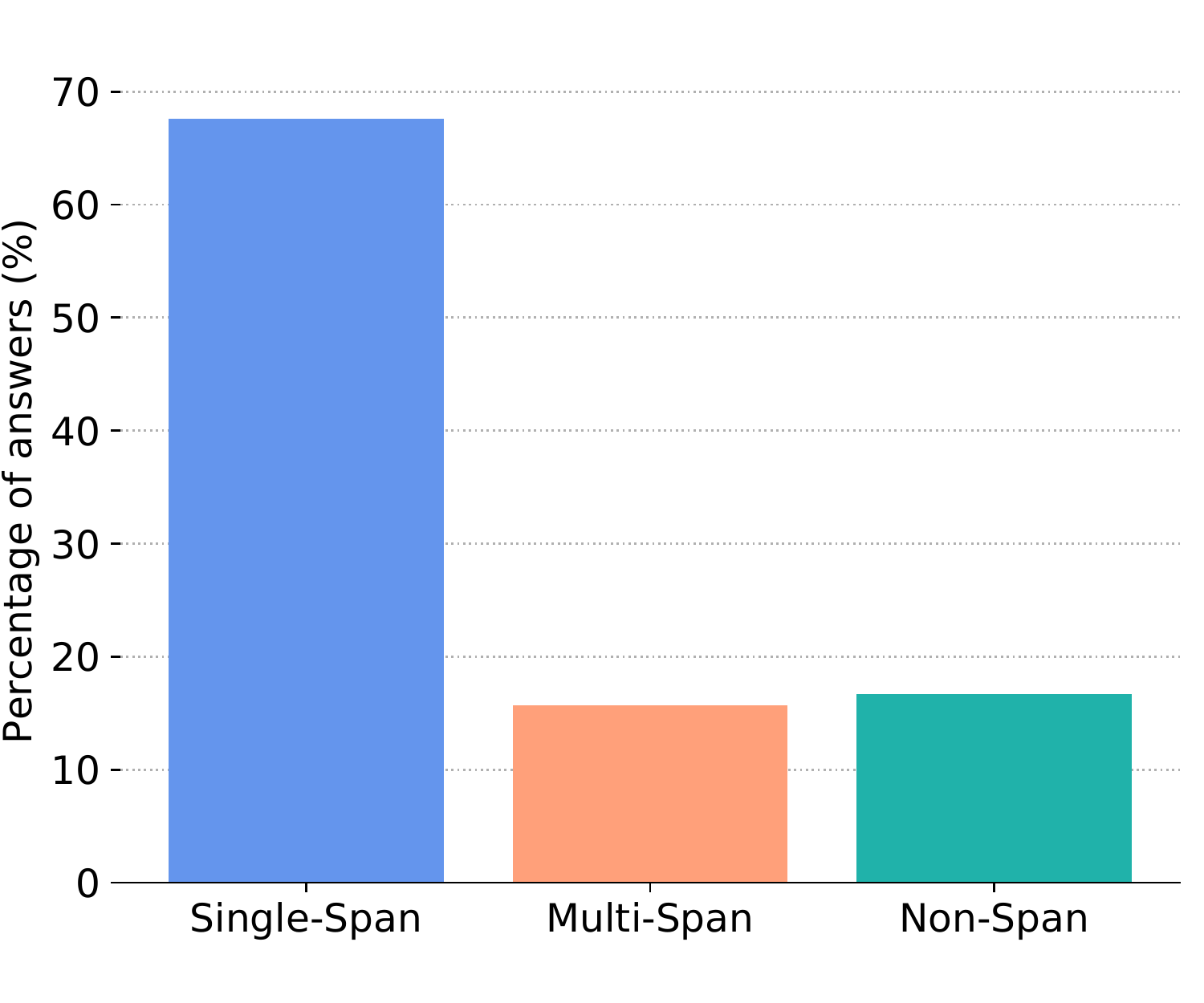}
    }    
    \caption{Distribution of bounding box categories, reasoning types, numerical operations, and answer types in the test set.}
    \label{fig:distributions}
\end{figure}

\subsubsection{Single-hop QA creation.}
We asked the workers to create 12,466 QA pairs by selecting a single slide
image from a slide deck. The selected slide can be used as evidence to tell whether a system arrived at the right answer for the right reasons. 
We encouraged questions that needed numerical reasoning, including operations of arithmetic expressions with $\{+, -, /, *\}$, counting, and comparisons. Additionally, the workers avoided creating questions that (i) contained selected page numbers; (ii) required external knowledge; (iii) were common to all of the slides (e.g., ``What is the title?").

\subsubsection{Multi-hop questions creation.}
We created 2,018 QA pairs for multi-hop reasoning by editing the single-hop questions created in %the QA collection step. 
the previous step.
For example at the left of Figure~\ref{fig:example_dataset}, ``North" is replaced by the phrase ``the region with 70\% of journals". To this end, we first identified one or two bridge entities 
in the created questions, and the workers selected related slides as evidence that mentioned the identified ones. Then, the content of the selected slides was utilized to replace the entities in the created questions. The process of creating multi-hop questions by editing may produce unnatural questions, as mentioned in the ``Limitations" section, but is easily scalable. A similar approach was taken with MultiModalQA~\cite{talmor2021multimodalqa}, which requires multi-hop reasoning over text, tables, and images in Wikipedia.

\subsubsection{Arithmetic expression annotation.} 
%\citet{Mathew_2022_WACV} have pointed out that numerical reasoning, especially arithmetical reasoning, is challenging for document VQA models. To remedy this problem, 
We provided arithmetic expressions 
like ``30  - 28"  in which the final numerical answer can be arrived at with the four arithmetic operations. The interpretation of the answer generation process is important for creating explainable QA models.

\subsection{Statistics and Analysis}

SlideVQA contains 14,484 QA pairs from 2,619 
slide decks, consisting of 52,480 slide images annotated with 890,945 bounding boxes. We split the dataset into 10,617 questions for training, %1,652 questions in development, and 2,215 question in test splits, making sure that each deck appears in the same split. 
1,652 (2,215) questions for development (test), making sure that each deck appears in the same split. 
%We compare SlideVQA with related datasets in terms of ``Images" and ``Questions and Answers". 

%We split the SlideVQA dataset into 10617 questions and 1,919 decks (38,380 images) in train, 1,652 question and 300 decks (6,000 images) in dev, and 2,215 question and 400 decks (8,000 images) in test splits, such that each deck appears in only one set. 

\subsubsection{Images.}
SlideVQA provides the largest number of images covering broad range of topics among the datasets shown in~Table~\ref{tab:statistics_dataset}. 
%As shown in Table~\ref{tab:statistics_dataset}, SlideVQA provides the largest number of \textcolor{blue}{bounding box annotations} among the datasets. Specifically, %\textcolor{blue}{the number of images is three times that of DocCVQA}, and 
Moreover, SlideVQA provides the largest number of bounding box annotations, where the number of the annotations in SlideVQA is 14.7 times that of VisualMRC. Figure~\ref{fig:distributions}a shows the distribution of bounding boxes broken down into nine categories, which cover all classes, including visually related ones (Image and Figure), unlike DocVQA and DocCVQA. To analyze the OCR tokens, we extracted the text shown in the images by using the Google Cloud
Vision API\footnote{https://cloud.google.com/vision}.  
As a result, 
the number of OCR tokens the system should consider simultaneously is larger (1488.88 tokens) than those of single-image document VQA datasets; the largest dataset (InfographicVQA) has 217.89 tokens. 

\begin{figure}[t!]
    \centering
    \includegraphics[width=.5\textwidth]{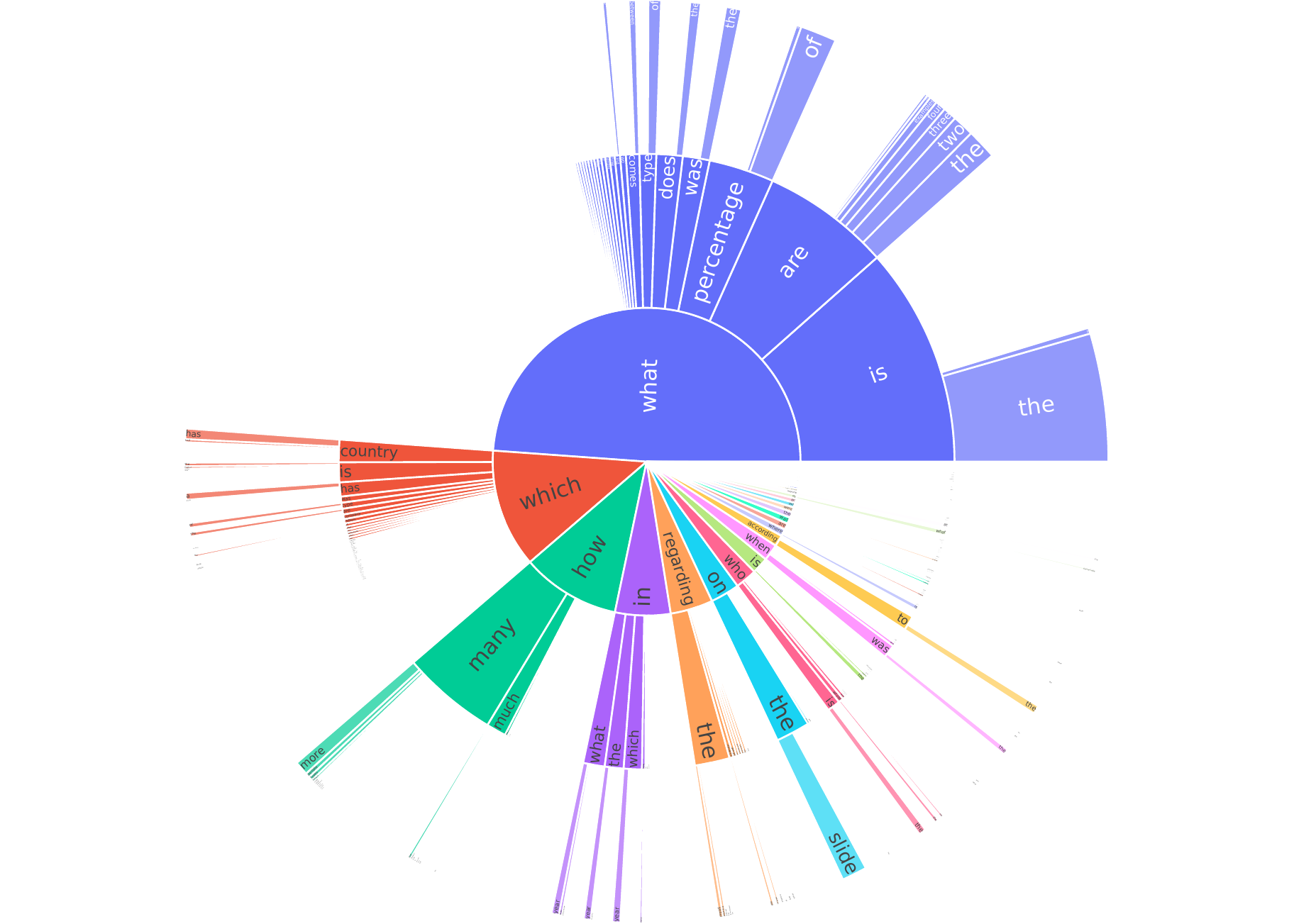}
    \caption{Distribution of the first three words of the questions.}
    \label{fig:sunburst}
\end{figure}

\begin{figure*}
    \centering
    \includegraphics[width=0.9\textwidth]{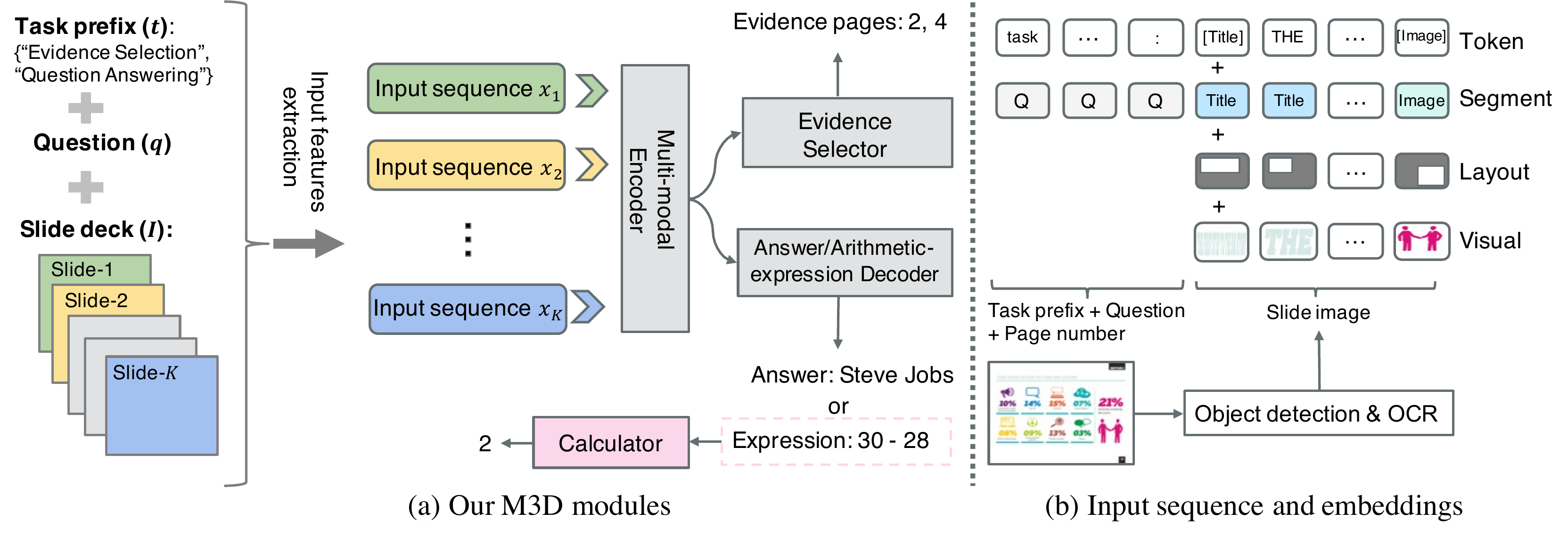}
    \caption{(a) Our encoder-decoder model architecture and (b) input representations. Given a question with a task prefix and a slide deck, the model outputs a corresponding answer/arithmetic-expression and evidence pages. The calculator outputs the final answer to calculate the generated arithmetic expression.}
    \label{fig:proposed_model}
\end{figure*}

\subsubsection{Questions and answers.}
As shown in Table~\ref{tab:statistics_dataset}, SlideVQA 
%is the largest document VQA dataset that requires
requires complex reasoning %over %\textcolor{blue}{multi-document images},
%\textcolor{magenta}{multiple images},
including single/multi-hop, and numerical reasoning. 
%We also annotate each question-answer pair with corresponding answer, reasoning, and numerical operation types in test data. 
Figure~\ref{fig:distributions}b shows the diverse distribution of questions related to reasoning types. 49.3\% of the questions require multi-hop or numerical reasoning.
%, while most existing datasets required only single-hop reasoning. 
%%
Moreover, SlideVQA
%the first dataset to provide
provides annotations of arithmetic expressions to improve numerical reasoning. Figure \ref{fig:distributions}c shows the distribution of numerical operations. 25.5\% of the numerical questions require arithmetic operations, which current systems have particular difficulty answering. Figure~\ref{fig:distributions}d shows that multi-span and non-span account for 32.4\% of the answers, indicating systems also need to generate answers as well as extract multiple spans.

Figure \ref{fig:sunburst} shows the sunburst pattern of the first three words of the questions. ``In" and ``Regarding" are frequent first words because SlideVQA needs to search for evidence images from a slide deck, which is a special pattern in multi-text document QA~\cite{Yang0ZBCSM18}.

%This distribution is similar to InforgraphicVQA, but the percentage of arithmetic operation in InfographicVQA has the lowest. 
%The Non-Span setting is similar to InfographicVQA, but it provides only numerical values as the Non-Span answer. 
% Numerical questions often starts with ``What percentage" and ``How many", 
%The diversity in the question distribution implies the requirement of high-level understanding of textual and visual contents, as well as layout understanding. 

\section{Our Model}
Figure~\ref{fig:proposed_model} shows an overview of our model, called M3D
(\textbf{M}ulti-\textbf{M}odal \textbf{M}ulti-image \textbf{D}ocument VQA model).
We use Fusion-in-Decoder (FiD)~\cite{izacard2020leveraging}, which is a state-of-the-art multi-text encoder-decoder model, as our base model and initialize FiD with a pre-trained T5~\cite{RaffelSRLNMZLL20}.
We extend FiD to perform the end-to-end SlideVQA task (defined in \textsc{MainTask}) by (i) performing evidence selection and question answering tasks as a unified sequence-to-sequence format using multi-task learning, (ii) predicting arithmetic expressions as intermediate reasoning steps instead of generating answers directly to enhance numerical reasoning, and (iii) modifying the input sequence to learn the visual layout and content of the image.
%We extend Fusion-in-Decoder (FiD)~\cite{izacard2020leveraging}, which is a state-of-the-art encoder-decoder model initialized from a pre-trained T5~\cite{RaffelSRLNMZLL20}, by (i) performing evidence selection and question answering tasks as a unified sequence-to-sequence using multi-task learning, (ii) predicting arithmetic expressions as intermediate reasoning steps instead of generating answers directly to enhance numerical reasoning, and (iii) modifying the input sequence to learn the visual layout and content of the %document image.

\subsection{Multi-modal Task-Specific Input}
\subsubsection{Input token sequence.}
For each image $I_k$, we first use Faster-RCNN ~\cite{ren2015faster}, which was trained on SlideVQA, to extract $N$ semantic regions (bounding boxes) and their labels
%region 
(e.g., Title and Image). We parse the slide image for each extracted region $r$ by using an OCR engine and apply a sub-word tokenizer to obtain OCR tokens $\mathbf{W}^r_k = \{w^{r}_{k,1},\ldots, w^{r}_{k,n}\}$ and corresponding OCR bounding boxes. To jointly train the evidence selection and question answering tasks, we add different task prefixes $t \in$ \{\texttt{Evidence Selection}, \texttt{Question Answering}\} to the encoder input. Specifically, the input sequence is as follows:
\begin{equation}
\nonumber
x_k = (\texttt{task:} t \texttt{ question:} q \texttt{ page:} e_k \texttt{ context:} c_k),
\end{equation}
where the sequence concatenates each slide and page number pair ($c_k$, $e_k$) with the question $q$ and task prefix $t$. To tell the role of each region, we insert region labels \texttt{[R$^{r_i}_{k}$]}, corresponding to the region label of the $i$-th region $r_i$ in $k$-th page, %just 
before the OCR tokens $\mathbf{W}^{r_i}_{k}$ extracted in $r_i$:
\begin{equation}
\nonumber
c_k = 
( [{\rm \texttt{R}}^{r_1}_{k}], \mathbf{W}^{r_1}_{k}, [{\rm \texttt{R}}^{r_2}_{k}], \mathbf{W}^{r_2}_{k}, \dots,
[{\rm \texttt{R}}^{r_N}_{k}], \mathbf{W}^{r_N}_{k} )
\end{equation}

\subsubsection{Input embedding.}
Following LayoutT5~\cite{DBLP:conf/aaai/TanakaNY21}, the input embeddings $\mathbf{z}$ of the encoder are defined by utilizing multi-modal information, including token $\mathbf{z}^{{\rm token}}$, segment $\mathbf{z}^{{\rm seg}}$, layout $\mathbf{z}^{{\rm lay}}$, and visual embeddings $\mathbf{z}^{{\rm vis}}$ as follows:
\begin{equation}
\nonumber
\mathbf{z} = {\rm LN}(\mathbf{z}^{{\rm token}} + \mathbf{z}^{{\rm seg}} + \mathbf{z}^{{\rm lay}} + \mathbf{z}^{{\rm vis}})  \in \mathbb{R}^{L \times d},
\end{equation}
where LN is a layer normalization~\cite{BaKH16}, and $L$ and $d$ are the length of the input sequence and a hidden vector size, respectively. The segment embedding indicates which regions are included in the input sequence. The layout embedding denotes the encoded bounding box coordinates of the token within the image. We normalize all coordinates by the size of images and use embedding layers to embed x-axis and y-axis features separately. The visual embedding is the appearance feature of each region and the OCR bounding boxes, which were obtained from Faster-RCNN. Note that the layout and visual embeddings are set to zero vectors for the task prefix, question, and page number.

\subsection{Multi-modal Encoder-Decoder}
\subsubsection{Multi-modal encoder.} 
Our encoder is a stack of $m$ Transformer blocks, consisting of a self-attention layer and a fully-connected layer with residual connections. 
Following FiD~\cite{izacard2020leveraging}, all $K$ input sequences are encoded independently and then 
concatenated to form a unified input representation. Formally, we transform each input sequence $x_k$ into $\mathbf{x}_k \in \mathbb{R}^{L \times d}$ and concatenate them into $\mathbf{X} \in \mathbb{R}^{K \times L \times d}$.

\subsubsection{Answer/Arithmetic-expression decoder.}
Our decoder is another stack of $m$ Transformer blocks similar to the multi-modal encoder, where each block has an additional layer of cross-attention between the output sequence and $\mathbf{X}$. The answer decoder is modeled
as a conditional generation $p_\theta(y|\mathbf{X})$, %given $\mathbf{X}$ consuming the unified input representation $\mathbf{X}$,
where $\theta$ represents the set of all model parameters. To allow the model to perform numerical reasoning, we train the system to predict annotated arithmetic expressions $y'$ (e.g., ``$30 - 28$") instead of numeric values $y$ (e.g., ``$2$") by modeling $p_\theta(y'|\mathbf{X})$. 
%To inform the model whether to generate answer tokens or the expression, 
During inference, the model itself decides whether numerical reasoning is required or not for each question by predicting an indicator token \texttt{Answer:} or \texttt{Expression:} at the beginning of the output sequence.

\subsubsection{Evidence selector.}
The selector shares the weights and the architecture of the answer/arithmetic-expression decoder. Instead of only modeling answer generation, we devise a simple method to train evidence selection in a unified sequence. Specifically, we define the output sequence as $\hat{\mathbf{I}}_{\text{pages}}$ $=$ (\texttt{Evidence pages:} $\hat{e}_1$, $\ldots$, $\hat{e}_{K'}$), where each $\hat{e}$ is the page number of the selected slide. 

\subsubsection{Training and inference.}
Our model is trained by minimizing the weighted sum of two losses $\mathcal{L} = \mathcal{L}_{\text{dec}} + \mathcal{L}_{\text{sel}}$, where $\mathcal{L}_{\text{dec}}$ and $\mathcal{L}_{\text{sel}}$ are the negative log-likelihood between the ground-truth and the prediction regarding the decoder and selector, respectively. During inference, we obtain the final prediction to post-process the decoded sequence by removing the task indicator. If an arithmetic expression is generated (i.e., \texttt{Expression:} is generated), we use a calculator to obtain the final results. 

\section{Experiments}

\begin{table*}[t!]
\small
\tabcolsep=3pt
    \begin{minipage}{0.32\textwidth}
        \begin{center}
            \vspace{1.15cm}
            \scalebox{0.9}{
            \begin{tabular}{l|c|cccc} 
                \toprule
                & & \multicolumn{2}{c}{Dev} & \multicolumn{2}{c}{Test} \\ 
                Model & Modal & JEM & JF1 & JEM & JF1   \\ 
                \midrule
                PreasM & T & 30.2 & 38.2 & 23.4 & 34.7 \\
                T5 & T & 30.0 & 38.0 & 22.6 & 34.2 \\  
                T5 + $\mathbf{z}^{{\rm lay}}$ & TL & 30.9 & 39.5 & 23.6 & 35.7   \\ 
                LayoutT5 & TLV & 31.7 & 39.9 & 24.3 & 36.1 \\ 
                LayoutLMv2\dag & TLV & 22.8 & 30.8 & 16.5 & 26.5 \\
                \midrule
                M3D & TLV & \textbf{36.2} & \textbf{42.8} & \textbf{28.0} & \textbf{37.3} \\   
                M3D$_{\texttt{GT}}$ & TLV & 44.6 & 50.4 & 35.4 & 44.7 \\  
                \midrule 
                Human & -- & -- & -- & 88.6 & 91.9 \\
                \bottomrule
            \end{tabular}
            }
        \end{center}
        \centering
        \text{(a) Performance of main task.}
    \end{minipage}
    %\hspace{0.5em}
    \begin{minipage}{0.34\textwidth}
        \vspace{0.36cm}
        \begin{center}
            \scalebox{0.9}{
            \begin{tabular}{l|c|cccc} 
                \toprule
                & & \multicolumn{2}{c}{Dev} & \multicolumn{2}{c}{Test} \\ 
                Model & Modal & EM & F1 & EM & F1   \\         
                \midrule
                BM25 & T & 40.1 & 46.0 & 35.9 & 47.5  \\
                \midrule
                CLIP$_{\texttt{zero}}$ & V & 33.0 & 34.8 & 30.6 & 34.4 \\
                CLIP & V & 40.6 & 43.0 & 39.3 & 43.5 \\
                BERT & T & 60.9 & 74.4 & 50.3 & 69.2 \\ 
                BERT + $\mathbf{z}^{{\rm lay}}$ & TL &  61.4 & 75.2 & 52.7 & 71.0 \\
                LayoutLM & TL & 51.0 & 63.7 & 42.0 & 59.9 \\
                LayoutLMv2 & TLV & 63.3 & 77.1 & 51.7 & 71.5 \\  
                H-LayoutLMv2 & TLV & 81.1 & \textbf{89.5} & 69.8 & \textbf{85.6} \\
                \midrule
                M3D & TLV &  \textbf{83.1} & 87.7 & \textbf{75.0} & 83.8   \\
                \midrule 
                Human & -- & -- & -- & 97.7 & 98.0  \\ 
                \bottomrule 
            \end{tabular}
            }
            %\label{tab:select_predicted}
        \end{center}
        \centering
        \text{(b) Performance of evidence selection task.}
    \end{minipage}  
    \begin{minipage}{0.3\textwidth}
        \begin{center}
           \scalebox{0.9}{
            \begin{tabular}{l|c|cccc} 
                \toprule
                & & \multicolumn{2}{c}{Dev} & \multicolumn{2}{c}{Test} \\ 
                Model & Modal & EM & F1 & EM & F1   \\ 
                \midrule
                Q-only & -- & 9.4 & 11.4 & 10.7 & 13.5  \\ 
                UniVL & V & 8.8 & 12.1 & 10.6 & 14.1 \\ 
                \midrule 
                PreasM & T & 36.3 & 41.9 & 30.7 & 38.2 \\
                T5  & T & 35.2 & 41.3 & 29.3 & 37.9 \\  
                T5 + $\mathbf{z}^{{\rm lay}}$ & TL &  36.9 & 43.2 & 31.0 & 39.7  \\ 
                LayoutT5  & TLV & 38.9 & 44.8 & 31.7 & 39.9 \\ 
                LayoutLMv2\dag & TLV & 26.5 & 33.4 & 21.4 & 29.3 \\
                \midrule
                FiD  & T & 37.6 & 42.9 & 30.4 & 38.9 \\
                FiD + $\mathbf{z}^{{\rm lay}}$ & TL & 38.1 & 43.3 & 30.6 & 38.9 \\ 
                M3D & TLV & \textbf{41.3} & \textbf{47.1} & \textbf{33.5} & \textbf{41.7} \\
                \midrule 
                Human & -- & -- & -- & 89.8 & 93.0  \\
                \bottomrule
            \end{tabular}
            }
        \end{center}
        \centering
        \text{(c) Performance of question answering task.}
    \end{minipage} 
    %\hspace{0.5em}
    \caption{Performance of SlideVQA tasks. ``T/L/V'' denotes the ``text/layout/visual" modality of images. \dag denotes the extractive approach. The pipeline models answer the question based on the top-3 evidences obtained by H-LayoutLMv2. M3D$_{\texttt{GT}}$ knows the ground-truth evidence.
    %slides. 
    + $\mathbf{z}^{{\rm lay}}$ denotes addition of the layout embedding to the input embeddings. LayoutLM was not pre-trained in any matching task (e.g., text-image matching). CLIP$_{\texttt{zero}}$ denotes CLIP without fine-tuning.
    }
    \label{tab:main}
\end{table*}

\subsection{Experimental Setup}
We conducted experiments on the SlideVQA task, evidence selection task, and question answering task respectively defined in \textsc{MainTask}, \textsc{Subtasks} \ref{prob:es} and \ref{prob:qa}. 

\subsubsection{Main task baselines.} 
We mainly evaluated pipeline models as baselines, consisting of evidence selection that produces top-3 evidences and question answering that takes the selection results as input. Here, we introduced a hierarchical LayoutLMv2 (H-LayoutLMv2) inspired by~\cite{tu2020select,xu2020layoutlmv2}, which encodes all slides simultaneously by using another Transformer layer, as the evidence selector. It achieved 96.0\% on Recall@3 on the test set.
We used three generative QA models: a textual model \textbf{T5}~\cite{RaffelSRLNMZLL20}, a numerical and multi-hop model \textbf{PreasM}~\cite{yoran-etal-2022-turning}, and a document VQA model \textbf{LayoutT5}~\cite{DBLP:conf/aaai/TanakaNY21}. 
We also used an extractive document VQA model \textbf{LayoutLMv2} to predict the single span.

\subsubsection{Evidence selection baselines.}

We also evaluated the evidence selection task alone.
%We used several %text retrieval models as baselines.
%\textbf{BM25}~\cite{robertson2009probabilistic} uses search engines to estimate the relevance of slide texts
%documents 
%to a given question, and we used the top-1 slide
%document 
%as the prediction. 
\textbf{BM25}~\cite{robertson2009probabilistic} is a non-neural retrieval framework to estimate the relevance of texts to a search query.
%, and we used the top-1 slide as the prediction.} 
%Regarding
For the neural models,
\textbf{CLIP}~\cite{radford2021learning} encodes the question and each image to predict the highest similar pair. BM25 and CLIP used the top-1 slide as the prediction.
\textbf{BERT}~\cite{DevlinCLT19} is a pre-trained language model which only uses text information with the Transformer architecture. \textbf{LayoutLM}~\cite{XuLCHWZ20} incorporates layout information into the input embeddings of BERT. \textbf{LayoutLMv2} includes image features produced by a CNN backbone in input embeddings. To model the interactions between the %documents, 
slides,
%we introduced a hierarchical LayoutLMv2 (\textbf{H-LayoutLMv2}) inspired by~\cite{tu2020select}, which encodes all slides
%documents simultaneously by using another Transformer layer. 
we used \textbf{H-LayoutLMv2} described in the previous section. For neural evidence selection baselines (except for CLIP), we use a hidden state of \texttt{[CLS]} in the last layer to feed into an MLP classifier with a sigmoid activation. Evidence is selected if its confidence of binary classification is above the optimal value on the development set.

%In all of the above neural baselines (except for CLIP), we used a hidden state of \texttt{[CLS]} in the last layer to feed into a MLP classifier with a sigmoid activation. Evidence is selected if its confidence of binary classification is above the optimal value on the development set. %To test the performance of humans, we asked another six workers to select slide images relevant to the question (\textbf{Human}).

To evaluate the effectiveness of our generative evidence selection module, we introduced \textbf{BinaryClass} as a classification baseline, which uses a two-layer MLP classifier with a sigmoid activation on top of each encoder representation at the start-of-sequence. We also introduced a generative baseline, \textbf{ChainGen}, which generates a sequence of selected slide page numbers before the answer~\cite{wei2022chain}.

\subsubsection{Question answering baselines.} 
In addition to the pipeline models,
we developed \textbf{Q-only}, which takes only the question into T5. %~\cite{RaffelSRLNMZLL20}, to check the data bias. 
We also used 
a VideoQA model \textbf{UniVL}~\cite{Luo2020UniVL} 
%as a state-of-the-art model of the VideoQA task~\cite{le-hoi-2020-video} 
that can take all of the slide images as input.
%as an end-to-end approach. 
Furthermore, we evaluated our base model \textbf{FiD}~\cite{izacard2020leveraging}. %as a QA baseline.

\iffalse
We developed \textbf{Q-only}, which takes only 
the question into T5, to check the data bias. Additionally, we selected \textbf{UniVL}~\cite{Luo2020UniVL} as a state-of-the-art model of the VideoQA task~\cite{le-hoi-2020-video}.
We also compared our model with the two-step pipeline models consisting of the evidence selector and the answer generator.
Specifically, we used H-LayoutLMv2 as the evidence selector to produce top-3 evidences, which achieved 96.0\% on Recall@3 on the test set. Then, we took the selection results into a state-of-the-art textual QA model \textbf{T5}, numerical and multi-hop QA model \textbf{PreasM}~\cite{yoran-etal-2022-turning}, and document VQA model \textbf{LayoutT5}~\cite{DBLP:conf/aaai/TanakaNY21}. We implemented \textbf{LayoutLMv2} as an extraction-based model to predict the single span. Our base model for the end-to-end approach was \textbf{FiD}~\cite{izacard2020leveraging}.
\fi

\subsubsection{Human performance.} We asked six crowdworkers (not among those recruited to collect our dataset) to select slide images relevant to the question and answer the question.

\subsubsection{Evaluation metrics.}
Following HotpotQA~\cite{Yang0ZBCSM18}, we used exact match (EM) and F1 on each question answering and evidence selection task and also used Joint EM (JEM) and Joint F1 (JF1) to evaluate both tasks. These joint metrics penalize models that perform poorly on either task and assess the accuracy and explainability of the question answering models.

\subsection{Implementation Details}
We implemented all of the models in PyTorch and experimented on eight Tesla V100 32GB GPUs. The size of CLIP was \texttt{Large} and the size of the other models was \texttt{Base}. We fine-tuned the models using AdamW~\cite{loshchilov2017decoupled} with a learning rate of 5e-5 and a dropout rate of 10\%, and we linearly warmed up the learning rate over 1000 steps. The batch size was set to 32. We evaluated models every 500 steps and selected the best one on the development set on the basis of the loss. We used a maximum length of 200 tokens for each input sequence of M3D, and set the maximum target sequence length to 50. We trained Faster-RCNN~\cite{ren2015faster} with a ResNet-101~\cite{HeZRS16} backbone by using stochastic gradient descent (SGD)~\cite{ruder2016overview} with a learning rate of 1e-3 and batch size of one. Standard anchor scales of [8, 16, 32] and anchor ratios of [0.5, 1.0, 2.0] were used. For the VideoQA baseline, we created a new video at a rate of five frames per second. We used the Google Cloud Vision API to extract text and bounding boxes from images. When the OCR word is tokenized into sub-word tokens, the bounding box coordinates of a sub-word token are the same as those of its whole word.

\subsection{Experimental Results and Analysis}
\subsubsection{Does our model outperform the baselines?}
Table~\ref{tab:main} summarizes the results of the main tasks. As shown in Table~\ref{tab:main}a, M3D outperformed the baselines on joint EM/F1, where the metrics evaluate the consistency between the predicted evidence and answers. For the evidence selection task, Table~\ref{tab:main}b shows that H-LayoutLMv2 and M3D performed better than the baselines. This indicates that modeling the interaction between multiple slides
simultaneously is needed to improve performance. For the QA task, Table~\ref{tab:main}c shows that M3D outperformed the pipeline methods
%, which used the candidates of the evidence obtained by H-LayoutLMv2, 
in all metrics. 
%\textcolor{blue}{We hypothesize that the pipeline models generated answers from only the higher-ranked slides as evidence, while M3D can generate answers from all of the slides, including the lower-ranked ones.}
Our end-to-end M3D model is better at ignoring the slides irrelevant to the question than
the answer generator in the pipeline methods that strongly depend on the slides narrowed down by the evidence selector. 
However, M3D$_{\texttt{GT}}$ in Table~\ref{tab:main}a achieved a significant improvement by knowing the ground-truth slides. There is room for improving the correctness of evidence selection. 
%Additionally, 

%; this indicates that 
%\subsubsection{Which is the effective QA method, extraction or generation?}
%Table~\ref{tab:main}a and Table~\ref{tab:main}c show that the generation-based model of LayoutT5 significantly outperformed the extractive-based approach of LayoutLMv2. This result is inline with observation in DROP dataset~\cite{dua-etal-2019-drop} that has multi-type answers~\cite{geva-etal-2020-injecting}.

\begin{figure}[t!]
    \centering
    \includegraphics[width=.47\textwidth]{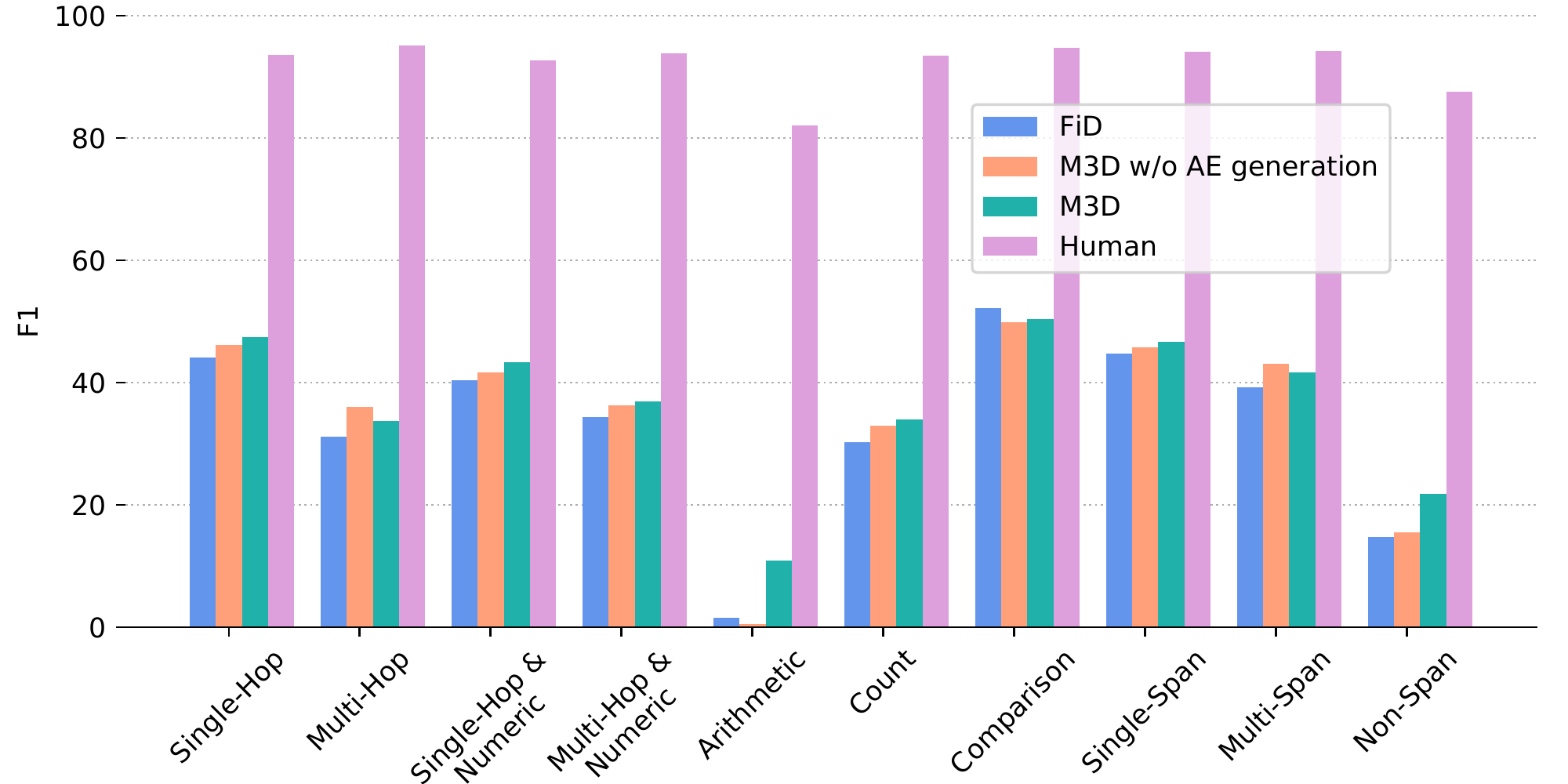}
    \caption{Performance of models and humans on the answer types, reasoning types and numerical operation types in the test set. AE stands for ``arithmetic expression''.}
    \label{fig:compare}
\end{figure}

\subsubsection{What are the characteristics of our dataset?}
Table~\ref{tab:main} shows that adding modality information tended to improve performance in all tasks. This demonstrates that SlideVQA requires methods to have the ability to jointly understand the text, layout, and visual modalities of documents. As shown in Table~\ref{tab:main}c, Q-only had the lowest performance, showing that the systems could not answer the question without reading documents in the SlideVQA task. 
Additionally, UniVL has a comparative result to Q-only, 
%indicating that the VideoQA model failed to read text in images}.
indicating that SlideVQA requires different abilities from VideoQA~\cite{le-hoi-2020-video}, especially the ability to read texts in images.
Tables~\ref{tab:main}a and \ref{tab:main}c show that LayoutT5, a generative model, significantly outperformed LayoutLMv2, an extractive approach. This result is inline with observations on the DROP dataset~\cite{dua-etal-2019-drop}, which also has non-span answers~\cite{geva-etal-2020-injecting}. Additionally, all of the models performed all of the tasks significantly worse than humans. To be specific, Figure~\ref{fig:compare} illustrates that (i) better multi-hop reasoning over multiple %document images 
images is needed and (ii) non-span answers to questions involving arithmetic operations have to be improved.

%\textcolor{blue}{TODO}

\begin{table}[t!]
    \centering
        \scalebox{0.83}{
    \begin{tabular}{lcccccc} 
        \toprule
        & \multicolumn{2}{c}{Main} & \multicolumn{2}{c}{Select} & \multicolumn{2}{c}{QA}  \\ 
        Model & JEM & JF1 & EM & F1  & EM & F1  \\ 
        \midrule
        M3D & \textbf{36.2} & \textbf{42.8} & \textbf{83.1} & \textbf{87.7} & \textbf{41.3} & \textbf{47.1} \\ \midrule
        w/o AE generation & 35.7 & 42.3 & 82.9 & 87.7 & 40.5 & 46.3   \\  
        w/o Evidence selection & -- & -- & -- & -- & 40.6 & 46.4  \\ 
        w/o Layout features & 35.1 & 42.0 & 82.4 & 87.1 & 40.3 & 46.3  \\  
        w/o Visual features & 34.2 & 40.9 & 81.5 & 86.3 & 39.0 & 44.9  \\ 
        w/o Text features & 1.0 & 1.5 & 8.4 & 9.8 & 9.8 & 12.0 \\ 
        \bottomrule
    \end{tabular}
    }
    \caption{Ablation study of M3D on dev set.}
    \label{tab:ablation}
\end{table}

\begin{table}[t!]
    \centering
        \scalebox{0.83}{
    \begin{tabular}{lcccccc} 
        \toprule
         & \multicolumn{2}{c}{Main} & \multicolumn{2}{c}{Select} & \multicolumn{2}{c}{QA} \\ 
        Model & JEM & JF1  & EM & F1 & EM & F1 \\ 
        \midrule
        M3D backbone & -- & -- & -- & -- & 39.0 & 44.8  \\ \midrule
        + BinaryClass & 24.7 & 34.8 & 54.5 & 68.5 & 38.8 & 44.8  \\
        + ChainGen & 34.0 & 40.8 & 81.1 & 86.1 & 39.8 & 45.4  \\
        + MultiGen (\textbf{Ours}) & \textbf{35.7} & \textbf{42.3} & \textbf{82.9} & \textbf{87.7} & \textbf{40.5} & \textbf{46.3}  \\ 
        \bottomrule
    \end{tabular}
    }
    \caption{Performance comparison of different evidence selection methods on dev set.}
    \label{tab:qa_classification}
\end{table}

\subsubsection{Do our sub-modules improve performance?} 
Table~\ref{tab:ablation} lists the results of an ablation study. Here, performance consistently decreased as individual modules were removed from M3D. This indicates that each of the modules is effective. More precisely, the arithmetic expression (AE) generation was influential on the QA and Joint performance, meaning that predicting the arithmetic expression instead of the numerical value enhances the ability to generate answers with numerical reasoning. %\textcolor{blue}{In token of that}, 
As shown in Figure~\ref{fig:compare}, applying AE prediction increased F1 by a large margin (+10.4\%) in the arithmetic type.

\subsubsection{What are the effective evidence selection methods?} 
%To test the effectiveness of our evidence selection method, we compared it with classification and another generation method. We introduced Binary Class for classification baselines, which uses an MLP classifier with a sigmoid on top of each encoder representation of the start-of-sequence token. For the generation baseline, %we were inspired from chain-of-thought~\cite{wei2022chain} and 
%we implemented Chain Gen, which generates a sequence of selected slide pages before the answer~\cite{wei2022chain}.
%Comparing models with the same backbone using different modeling approaches, 
Table~\ref{tab:qa_classification} shows that our method, which generates the evidence selection and question answering results separately, obtained the highest performance.
%, compared to the classification-based method (Binary Class) based on encoder representations and the generation-based method (Chain Gen) that generates a concatenated sequence of the slide page numbers and the answer.
%\textcolor{blue}{This result might be highly influenced by the text-to-text pretraining task of T5, which is similar to our training task}. 
It seems that the generative methods (MultiGen and ChainGen) benefited from the text-to-text pre-training of T5 more than the classification-based method (BinaryClass). Our MultiGen decoder that separately trains evidence selection and question answering had the advantage of being easier to train than the ChainGen baseline decoder that trains the two tasks as a single sequence generation task.

\subsubsection{On which categories does the object detection model not work well?}
\begin{table}[t!]
    \centering
        \scalebox{0.85}{
    \begin{tabular}{lcc} \toprule
    Class & Dev AP & Test AP \\ \midrule 
    Title & 86.8 & 87.5  \\
    Page-text & 76.9 & 76.9  \\
    Obj-text & 29.5 & 33.4  \\
    Caption & 25.6 & 24.9 \\
    Other-text & 40.5 & 39.4 \\
    Image & 60.4 & 62.2 \\
    Diagram & 65.4 & 64.0  \\
    Figure & 74.1 & 68.8  \\
    Table & 67.0 & 65.6 \\ \bottomrule
    \end{tabular}
    }
    \caption{Object detection performance of Faster-RCNN broken down by bounding box categories. We set an intersection-over union (IoU) threshold to 0.5.}
    \label{tab:object_detection}
\end{table}
Table~\ref{tab:object_detection} lists the object detection performance of Faster-RCNN broken down by bounding box categories. These results show that detecting randomly placed and small boxes, such as Obj-text, is more difficult than mostly fixed and large boxes, such as Title. 

\begin{figure}[t!]
    \centering
    \includegraphics[width=.495\textwidth]{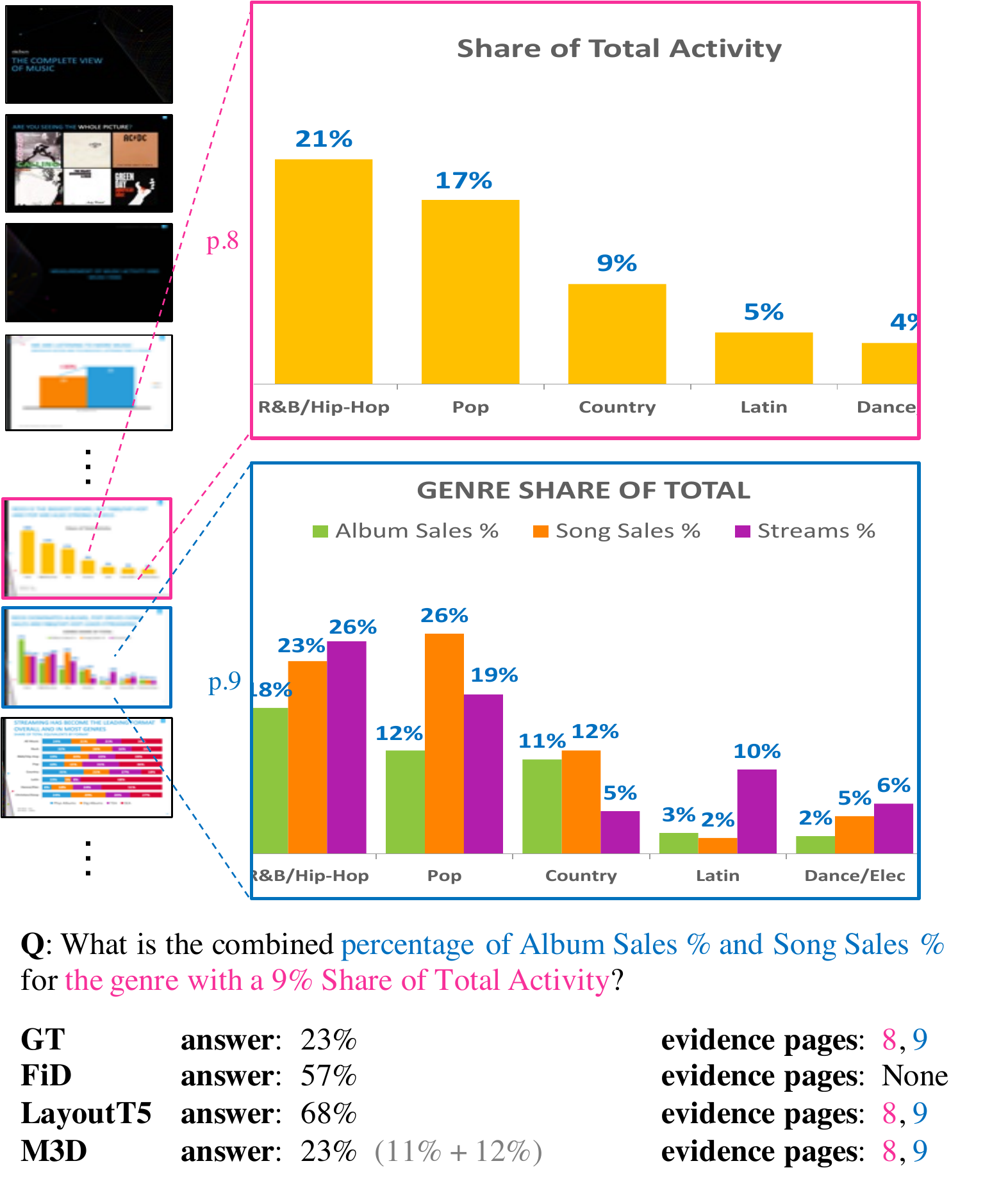}
    \caption{Qualitative example. GT denotes the ground-truth. ($\cdot$) means the generated arithmetic expression. The slide deck can be viewed at~\url{https://www.slideshare.net/musicbizassoc/nielsen-2015-music-biz-presentation-final}.}
\label{fig:qualitative}
\end{figure}

\subsubsection{Qualitative examples.}
%We demonstrate our model performance by visualizing a qualitative example in Figure~\ref{fig:qualitative}. 
Figure~\ref{fig:qualitative} demonstrates our model's performance by visualizing a qualitative example. This example needs multi-hop reasoning %over multiple documents 
and an answer involving an arithmetic operation. FiD gave an incorrect answer because it did not consider the visual layout of the slides.
%documents. 
Moreover, while LayoutT5 could not understand the process of getting numerical answers, M3D successfully extracted information (``11\%" and ``12\%") and generated the same answer as the ground-truth.  

\section{Discussion and Limitations}
% Although our proposed dataset and models provide important directions that enable systems to read diverse real-world documents, it has several limitations. 
%A limitation of our dataset is that \textcolor{blue}{a few} multi-hop questions created by editing would be different from questions humans ask to a system. 
SlideVQA is the largest document VQA benchmark that uses multiple %document 
images as input and requires multi-hop reasoning; its limitation is that 
the multi-hop questions created by editing are different from the questions humans might actually ask the system. We argue that developing models that  
can reason over multiple images is an important research direction, and therefore, we employed an editing method that guarantees multi-hop questions and easily extends the dataset size. 
%\textcolor{magenta}{As a result, we were able to create the largest document VQA dataset that takes multi-image input and requires multi-hop reasoning.}
%, which is the largest document VQA dataset takes multi-image input and requires multi-hop reasoning, 
%SlideVQA, \textcolor{magenta}{which is the largest document VQA dataset takes multi-image input and requires multi-hop reasoning}, provides an opportunity to develop and evaluate such models.
%Additionally, our collection method allows to guarantee multi-hop questions and to scale the dataset size easily. 
Also, our model uses cross-attention on all evidence candidates, which may cause a computational problem when there are a lot of input images (e.g., as in the open-domain QA setting like DocCVQA). To remedy this problem, 
we consider that models that train a two-stage selector that roughly narrows down candidates to a small number of images and then accurately selects evidence images and an answer generator in an end-to-end manner are promising~\cite{sachan-etal-2021-end,sachan2021endtoend}.

\section{Conclusion}
We introduced a new document VQA dataset, SlideVQA, focused on the task of understanding slide decks composed of multiple images. We also introduced a unified end-to-end model, M3D, that can perform evidence selection and question answering tasks and enhance numerical reasoning by generating arithmetic expressions. While our evaluation highlighted the promise of this approach, it also revealed a huge gap compared with human performance, and several challenges emerge from multi-hop reasoning
on multiple images and generating answers with arithmetic operations.
We believe that our dataset will contribute to the development of intelligent assistant agents that can comprehend diverse real-world documents.
\newpage
\bibliography{references}

\end{document}

%\linenumbers % camera readyで消す

\maketitle

\begin{figure}[t!]
    \centering
    \includegraphics[width=.44\textwidth]{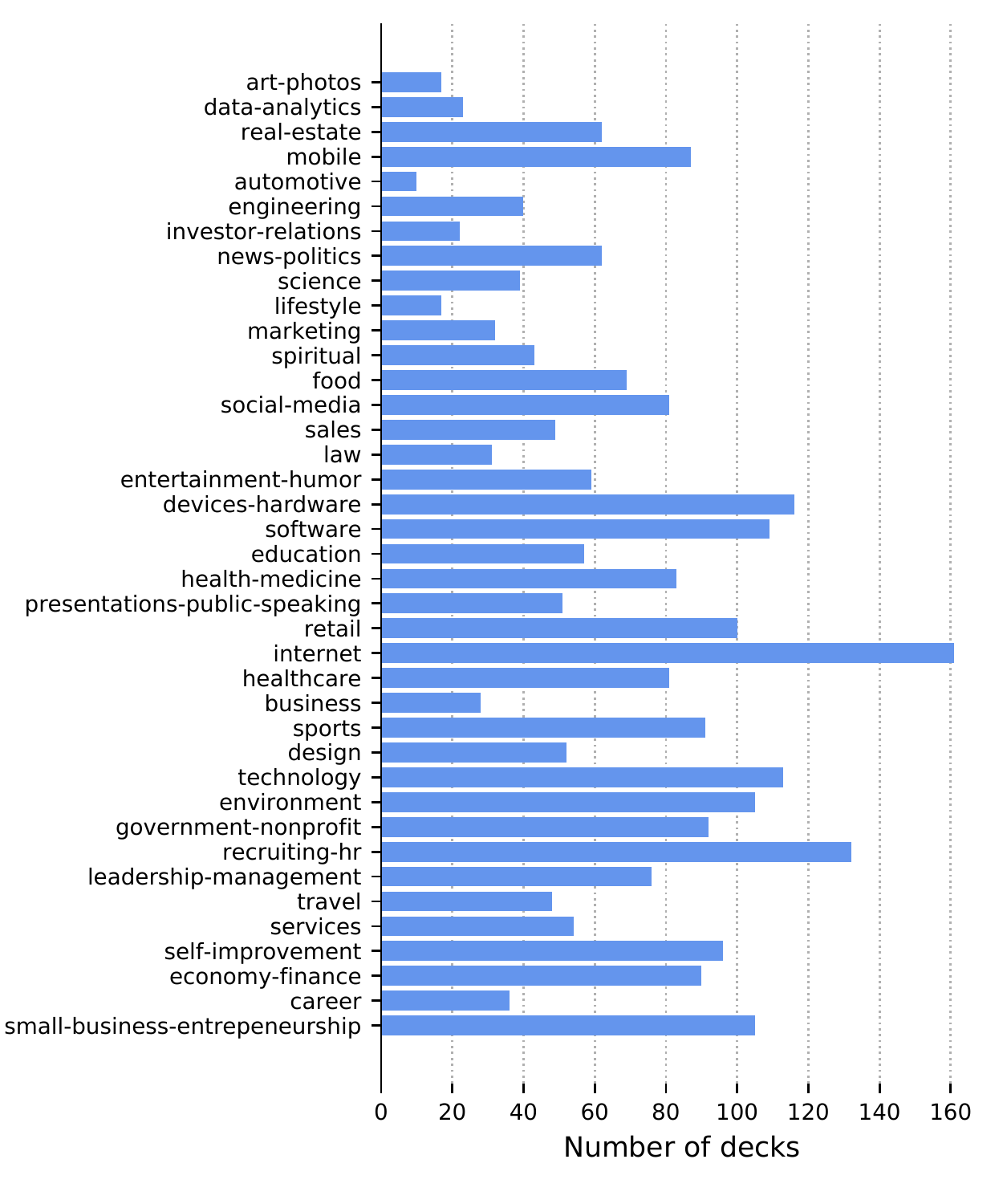}
    \caption{Categories of collected slide decks.}
    \label{fig:category}
\end{figure}

\section{Data Collection and Analysis}
\subsection{Slide decks collection.}
Figure~\ref{fig:category} shows the categories of the slide decks collected from slideshare. SlideVQA covers 39 categories.

\begin{figure}[t!]
    \centering
    \includegraphics[width=.44\textwidth]{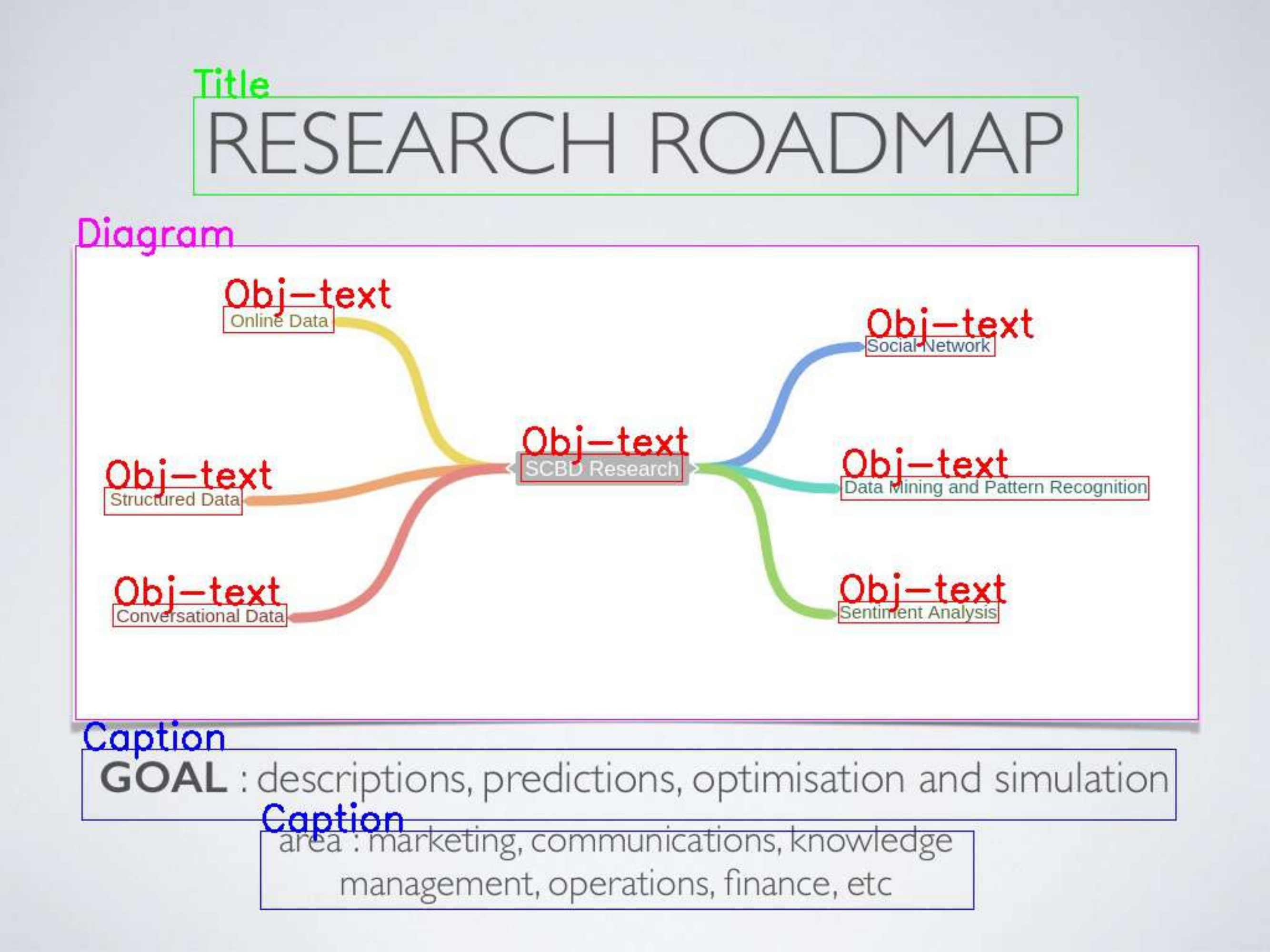}
    \caption{Example of collected bounding boxes. Colored boxes and words were annotated by crwodworkers.}
    \label{fig:bbox}
\end{figure}

\subsection{Bounding box annotations.}
We defined the bounding box categories inspired by the previous work~\cite{haurilet2019spase} as follows:
\begin{itemize}
    \item \textbf{Title}: presentation title, slide title
    \item \textbf{Page-text}: text in slide, bullet-point text list, text list
    \item \textbf{Obj-text}: text that is inside a figure, image, diagram or table
    \item \textbf{Caption}: description of figure, image, diagram, or table
    \item \textbf{Other-text}: footnote, date, affiliation, slide-number, programming code, URL
    \item \textbf{Diagram}: a graphical representation of data, a process, an idea, etc.
    \item \textbf{Table}: data arranged in rows and columns
    \item \textbf{Image}: drawing, logo, map, screenshot, realistic image
    \item \textbf{Figure}: graph with data points and coordinates
\end{itemize}
As shown in Figure~\ref{fig:bbox}, SlideVQA provides densely annotated bounding boxes. 

\subsection{Starting words of the question.}

\begin{figure}[t!]
    \centering
    \includegraphics[width=.49\textwidth]{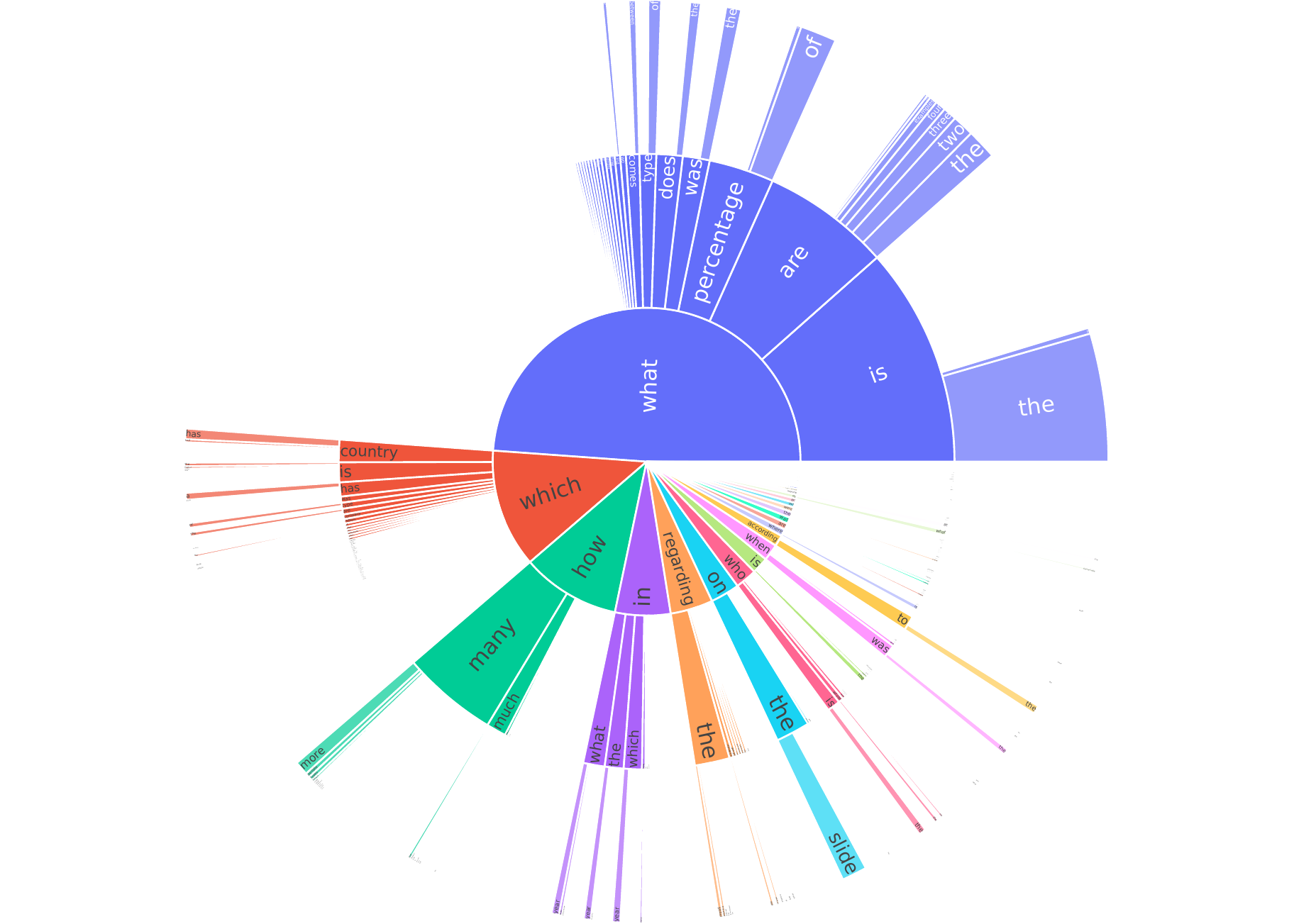}
    \caption{Distribution of the first three words of the questions.}
    \label{fig:sunburst}
\end{figure}

Figure \ref{fig:sunburst} shows the sunburst pattern of the first three words of the questions. ``In" and ``Regarding" are frequent first words because SlideVQA needs to search for evidence images from a slide deck, which is a special pattern in multi-text document QA~\cite{Yang0ZBCSM18}. 

\begin{figure}[t!]
    \centering
    \subfigure[Word cloud of questions.]{
    \includegraphics[width=.29\textwidth]{Images/question_cloud.pdf}
    }
    \hspace{1em}
    \subfigure[Word cloud of answers.]{
    \includegraphics[width=.29\textwidth]{Images/answer_cloud.pdf}
    }
    \hspace{1em}
    \subfigure[Word cloud of document images.]{
    \includegraphics[width=.29\textwidth]{Images/document_cloud.pdf}
    }
    \caption{Statistics of tokens in the question, answer, the document images. 
    }
    \label{fig:statistics}
\end{figure}
\subsection{Word clouds.}
Figure~\ref{fig:statistics} shows the word clouds of the questions, document images, and answers. It shows how diverse the vocabulary space is in SlideVQA. 

\begin{table}[t!]
    \centering
        \scalebox{0.8}{
    \begin{tabular}{lccccc} 
    \toprule
    Models   & lr & optimizer & bs & input length & epoch  \\ \midrule
    \textbf{Evidence selection} & & & & & \\
    CLIP & 1e-5 & Adam$^1$ & 32 & 76 & 5 \\
    BERT     & 5e-5 & Adam$^1$ & 8 & 250 & 3 \\
    BERT + $\mathbf{z}^{{\rm lay}}$ & 5e-5 & Adam$^1$ & 8 & 250 & 3\\
    LayoutLM & 2e-5 & Adam$^1$ & 8 & 250 & 3\\
    LayoutLMv2 & 5e-5 & Adam$^1$ & 8 & 250 & 3\\
    H-LayoutLMv2 & 5e-5 & Adam$^1$ & 8 & 250 ($\times$ 20) & 3 \\ \midrule
    \textbf{Question answering} & & & & & \\
    UniVL & 3e-5 & Adam$^1$ & 16 & 30 & 5 \\
    PreasM & 1e-4 & AdaFactor$^2$ & 160 & 512 & 20 \\
    T5 & 3e-5 & AdamW$^3$ & 32 & 512 & 5 \\
    T5 + $\mathbf{z}^{{\rm lay}}$ & 5e-5 & AdamW$^3$ & 32 & 512 & 5 \\
    LayoutT5 & 5e-5 & AdamW$^3$ & 32 & 512 & 5 \\
    LayoutLMv2 & 5e-5 & AdamW$^3$ & 40 & 512 & 20 \\
    FiD & 5e-5 & AdamW$^3$ & 32 & 190 ($\times$ 20) & 5 \\
    FiD + $\mathbf{z}^{{\rm lay}}$ & 5e-5 & AdamW$^3$ & 32 & 190 ($\times$ 20) & 5 \\ \bottomrule
    \end{tabular}
        }
    \caption{Hyperparameters of the baselines in SlideVQA. lr and bs denotes the learning rate and batch size, respectively.$^1$\cite{KingmaB15}; $^2$\cite{shazeer2018adafactor}; $^3$\cite{loshchilov2017decoupled}}
    \label{tab:hyperparameters}
\end{table}

\section{Experiments}
\subsection{Experimental Setup}

\subsubsection{Baseline details.}
H-LayoutLMv2 first encodes each image and the question by using LayoutLMv2~\cite{xu2020layoutlmv2}, which accepts text, layout, and visual features as input. After encoding, we extract the hidden representations $\mathbf{h}$ of \texttt{[CLS]} in the last layer and then take one layer of the Transformer as the document encoder to obtain the representations $\mathbf{h}^{\text{doc}}$ that model the relationship between all the images. Finally, the hidden state of \texttt{[CLS]} is updated as follows:
\begin{equation}
\nonumber
\hat{\mathbf{h}} = \mathbf{h} + \mathbf{h}^{\text{doc}}
\end{equation}

For neural evidence selection baselines (except for CLIP), we use a hidden state of \texttt{[CLS]} in the last layer to feed into a MLP classifier with a sigmoid activation. Evidence is selected if its confidence of binary classification is above the optimal value on the development set.

LayoutLMv2 for question answering task needs to consider multiple images as input simultaneously. To this end, we created a new image by aligning the images horizontally. 

\subsubsection{Implementation details.}
Table~\ref{tab:hyperparameters} lists the hyperparameters of the baselines in SlideVQA.
We trained Fastre-RCNN~\cite{ren2015faster} with a ResNet-101~\cite{HeZRS16} backbone by using stochastic gradient descent (SGD)~\cite{ruder2016overview} with a learning rate of 1e-3 and batch size of one. Standard anchor scales of [8, 16, 32] and anchor ratios of [0.5, 1.0, 2.0] were used. For the VideoQA baseline, we created a new video at a rate of five frames per second. We used the Google Cloud Vision API\footnote{https://cloud.google.com/vision} to extract text from images. When the OCR word is tokenized into sub-word tokens, the bounding box coordinates of a sub-word token are the same as those of its whole word.

\subsection{Experimental Results and Analysis}

\subsubsection{Performance comparison with different OCR engines.}

\begin{table}[t!]
    \centering
        \scalebox{0.85}{
    \begin{tabular}{lcccccc} 
        \toprule
        & \multicolumn{2}{c}{Main}  & \multicolumn{2}{c}{Select} & \multicolumn{2}{c}{QA} \\ 
        OCR engine & JEM & JF1  & EM & F1 & EM & F1 \\ 
        \midrule
        Vision API & \textbf{36.2} & \textbf{42.8} & \textbf{83.1} & \textbf{87.7} & \textbf{41.3} & \textbf{47.1}  \\
        Tesseract & 22.5 & 28.3 & 69.6 & 74.7 & 28.3 & 34.0  \\
        \bottomrule
    \end{tabular}
    }
    \caption{M3D performance comparison with different OCR engines in the dev set.}
    \label{tab:ocr}
\end{table}
Table~\ref{tab:ocr} presents the results of M3D for the Vision API and Tesseract OCR engine. The differences in score are huge for all tasks and show a clear advantage for Vision API. The future direction includes that we will create models showing the robustness to variations in OCR quality.

\subsubsection{Performance of object detection.}
\begin{table}[t!]
    \centering
        \scalebox{0.9}{
    \begin{tabular}{lcc} \toprule
    Class & Dev AP & Test AP \\ \midrule 
    Title & 86.8 & 87.5  \\
    Page-text & 76.9 & 76.9  \\
    Obj-text & 29.5 & 33.4  \\
    Caption & 25.6 & 24.9 \\
    Other-text & 40.5 & 39.4 \\
    Image & 60.4 & 62.2 \\
    Diagram & 65.4 & 64.0  \\
    Figure & 74.1 & 68.8  \\
    Table & 67.0 & 65.6 \\ \bottomrule
    %mAP & 37.2 & 35.6 \\ \bottomrule 
    \end{tabular}
    }
    \caption{Object detection performance of Faster-RCNN broken down by bounding box categories. We set a intersection-over union (IoU) threshold to 0.5.}
    \label{tab:object_detection}
\end{table}
Table~\ref{tab:object_detection} lists the object detection performance of Faster-RCNN broken down by bounding box category. These results show that randomly placed and small boxes, such as Obj-text, are inferior to mostly fixed and large boxes, such as Title. 

\section{Reproducibility Checklist}

\subsection{Includes a conceptual outline and/or pseudocode description of AI methods introduced.}
Yes

\subsection{Clearly delineates statements that are opinions, hypothesis, and speculation from objective facts and results.}
Yes

\subsection{Provides well marked pedagogical references for less-familiare readers to gain background necessary to replicate the paper.}
Yes

\subsection{Does this paper rely on one or more datasets?}
Yes

\subsubsection{A motivation is given for why the experiments are conducted on the selected datasets.} 
Yes

\subsubsection{All novel datasets introduced in this paper are included in a data appendix.}
Yes

\subsubsection{All novel datasets introduced in this paper will be made publicly available upon publication of the paper with a license that allows free usage for research purposes.} 
Yes. Our dataset is based on images collected from slideshare. Following the user policy of slideshare, we will distribute download URL links instead of images. All other annotations will be available on our project page.

\subsubsection{All datasets drawn from the existing literature (potentially including authors’ own previously published work) are accompanied by appropriate citations.}
Yes

\subsubsection{All datasets drawn from the existing literature (potentially including authors’ own previously published work) are publicly available.} 
Yes

\subsubsection{All datasets that are not publicly available are described in detail, with explanation why publicly available alternatives are not scientifically satisficing.} 
NA

\bibliography{references}